%% file: paper.tex
\documentclass[onecolumn]{fairmeta}
\usepackage[most]{tcolorbox}

\usepackage{amsmath}
\usepackage{amssymb}
\usepackage[utf8]{inputenc} 
\usepackage[T1]{fontenc}    
\usepackage{hyperref}       
\usepackage{url}            
\usepackage{booktabs}       
\usepackage{amsfonts}       
\usepackage{nicefrac}       
\usepackage{microtype}      
\usepackage{xcolor}         
\usepackage{graphicx}
\newcommand{\m}[1]{{A2A}}
\newcommand{\p}[1]{}

\usepackage{wrapfig}

\title{Action-to-Action Flow Matching}

\author[1,*]{\href{https://jiajindou.github.io/}{\textcolor{black}{Jindou Jia}}}
\author[1,*]{\href{https://reagan1311.github.io/}{\textcolor{black}{Gen Li}}}
\author[1]{\href{https://xyc0212.github.io/}{\textcolor{black}{Xiangyu Chen}}}
\author[1]{\href{https://morpheus-an.github.io/}{\textcolor{black}{Tuo An}}}
\author[1]{Yuxuan Hu}
\author[1]{Jingliang Li}
\author[1]{Xinying Guo}
\author[1]{\href{https://marsyang.site/}{\textcolor{black}{Jianfei Yang}}}

\affiliation[1]{MARS Lab, Nanyang Technological University}

\abstract{
	Diffusion-based policies have recently achieved remarkable success in robotics by formulating action prediction as a conditional denoising process. However, the standard practice of sampling from random \textit{Gaussian} noise often requires multiple iterative steps to produce clean actions, leading to high inference latency that incurs a major bottleneck for real-time control. In this paper, we challenge the necessity of uninformed noise sampling and propose Action-to-Action flow matching (\m{}), a novel policy paradigm that shifts from random sampling to initialization informed by the previous proprioceptive action. Unlike existing methods that treat proprioceptive action feedback as static conditions, \m{} leverages historical proprioceptive sequences, embedding them into a high-dimensional latent space as the starting point for action generation. This design bypasses costly iterative denoising while effectively capturing the robot's physical dynamics and temporal continuity.
	Extensive experiments demonstrate that \m{} exhibits high training efficiency, fast inference speed, and improved generalization. Notably, \m{} enables high-quality action generation in as few as a single inference step, and exhibits superior robustness to visual perturbations and enhanced generalization to unseen configurations. Lastly, we also extend \m{} to video generation, demonstrating its broader versatility in temporal modeling. Project site: \url{https://lorenzo-0-0.github.io/A2A_Flow_Matching}.
}

\correspondence{\email{jindou.jia@ntu.edu.sg}, \email{jianfei.yang@ntu.edu.sg}}
\contribution[*]{Equal contribution}

\begin{document}
	
	\maketitle
	
	\begin{figure*}
		\centering
		\includegraphics[width=1\linewidth]{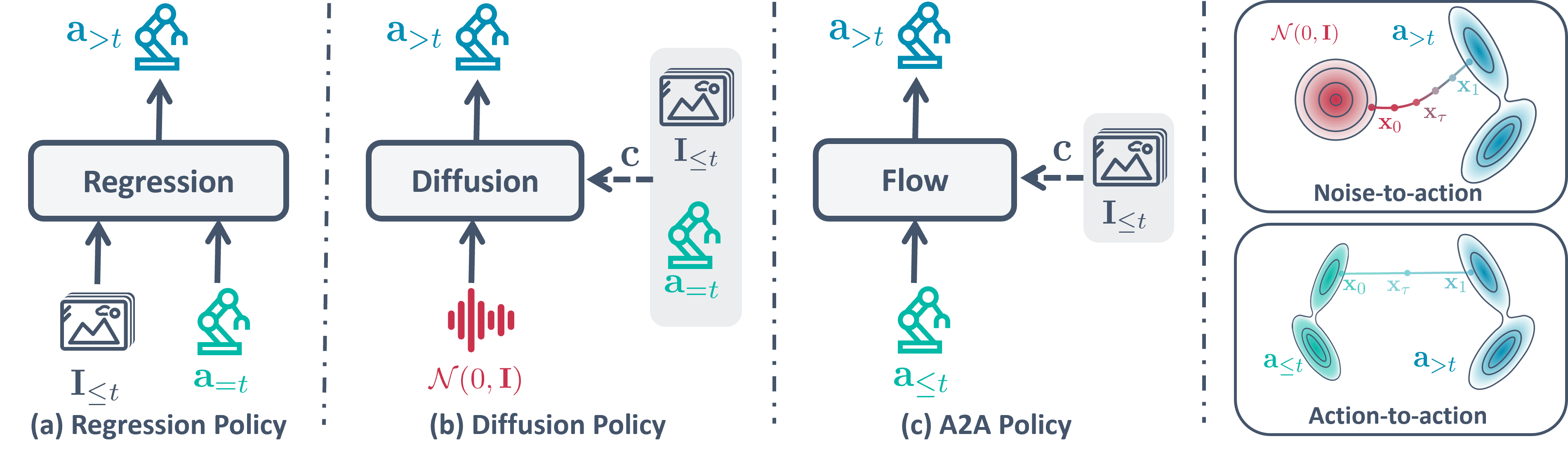}
		\caption{\textbf{Comparison of robotic policy paradigms.} (a) Regression Policy: Deterministic mapping from multi-modal inputs to actions. (b) Diffusion Policy: Generative modeling via iterative denoising from \textit{Gaussian} noise. (c) \m{} Policy: Informed action generation through a structured flow between historical and future actions. \textit{Action-to-action} allows for more efficient transport than \textit{noise-to-action}, enabling one-step flow mapping feasible even with a lightweight  MLP architecture.}
		\label{framework}
	\end{figure*}
	
	\input{sec/intro}
	\input{sec/related}
	
	\section{Action-to-action flow matching}
	
	\subsection{Flow matching}
	We adopt flow matching~\citep{flowmatching} as the algorithmic foundation, given its widespread adoption in the robotics field~\citep{black2024pi_0, intelligence2025pi, bjorck2025gr00t}. Flow matching provides a simulation-free training objective that learns to transform a source distribution $p_0 = \mathcal{N}(\mathbf{0}, \mathbf{I})$ into a target data distribution $p_1 = p_{\text{data}}$ over $\mathbb{R}^d$. Let $p_\tau: \mathbb{R}^d \to \mathbb{R}_{>0}$ denote a time-dependent probability density for $\tau \in [0, 1]$, defining a probability path that interpolates between $p_0$ and $p_1$. This path is generated by a time-dependent vector field $\mathbf{v}: [0, 1] \times \mathbb{R}^d \to \mathbb{R}^d$ via the ordinary differential equation (ODE)
	\begin{equation}
		\begin{aligned}
			\frac{d\mathbf{x}_\tau}{d\tau} = \mathbf{v}_\tau(\mathbf{x}_\tau), \qquad \mathbf{x}_0 \sim p_0,
		\end{aligned}
	\end{equation}
	where $\mathbf{x}_\tau \in \mathbb{R}^d$ denotes the state at flow time $\tau$.
	
	\begin{wrapfigure}{r}{0.5\textwidth}
		\centering
		\includegraphics[width=0.9\linewidth]{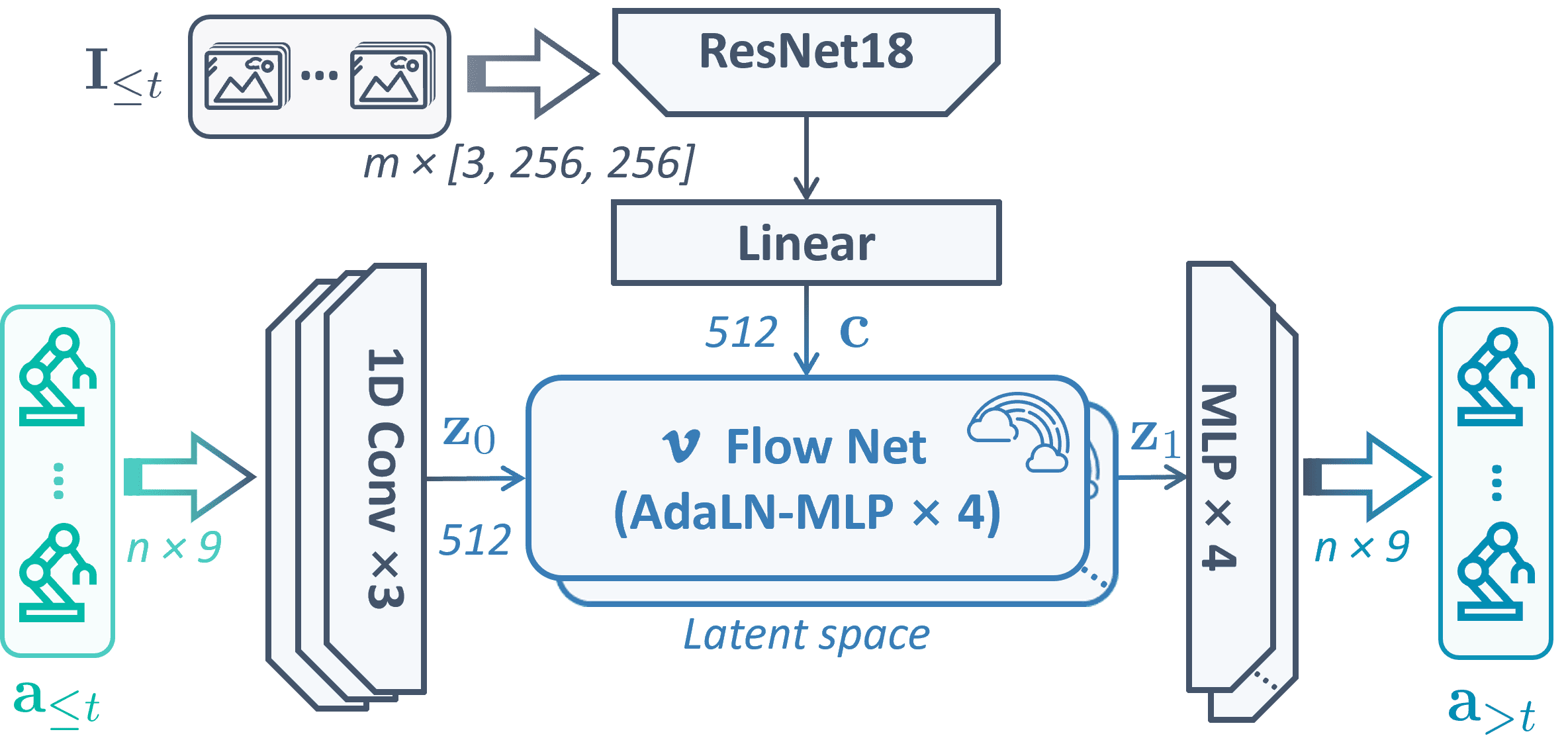}
		\caption{\textbf{Overview of \m{} architecture.} The framework consists of three main components. 1) A condition path that encodes visual observations using a {ResNet-18} backbone and a linear projector to generate a global condition $c$. 2) A source path that employs a CNN with a 5 kernel size to compress the $n$-frame history actions into a latent starting point $\mathbf{z}_0$. 3) A flow-based generation process. The flow net, built with {AdaLN-MLP} blocks, predicts the vector field to transport $\mathbf{z}_0$ to the target latent $\mathbf{z}_1$ within a unified 512-dimensional latent space. Finally, a residual MLP decoder transforms $\mathbf{z}_1$ into the future action sequence.}
		\label{pipeline}
	\end{wrapfigure}
	
	The optimal transport displacement map~\citep{flowmatching} is adopted, which is formalized as $\mathbf{x}_\tau = (1 - \tau)\mathbf{x}_0 + \tau \mathbf{x}_1$. The goal is to train a neural network $f_\theta(\mathbf{x}_\tau, \tau, \mathbf{c})$ parameterized by $\mathbf{\theta}$ to approximate the conditional vector field, where $\mathbf{c}$ represents the external conditioning (e.g., visual observations and proprioceptive states). The flow matching loss is defined as 
	\begin{equation}
		\begin{aligned}
			\mathcal{L}_{FM} = \mathbb{E}_{\tau \sim \mathcal{U}[0,1], \mathbf{x}_0\sim p_0, \mathbf{x}_1\sim p_1} \left\| f_\theta(\mathbf{x}_\tau, \tau, \mathbf{c}) - \mathbf{v}_\tau(\mathbf{x}_\tau) \right\|^2.
			\label{general_loss}
		\end{aligned}
	\end{equation}
	With the learned $f_\theta(\mathbf{x}_\tau, \tau, \mathbf{c})$, sampling in the inference phase is usually conducted via discretized \textit{Euler} integration.

	\subsection{Action to action flow}
	Different from previous generative-based polices that denoise from \textit{Gaussian} distributions \citep{ddpm, flowmatching, dp}, \m{} aims to learn a policy from a proprioceptive historical action space to a future action space, as depicted in Fig. \ref{framework}(c). The formulation of the action $\mathbf{a}$ is task-dependent and can accommodate various control modalities, including joint positions and end-effector states. We chose the joint angles in simulation and the end-effector states in experiments. Given historical actions $\mathbf{a}_{\leq t} = \{\mathbf{a}_{t-n+1}, \dots, \mathbf{a}_t\}$\footnote{To enhance closed-loop robustness, $\mathbf{a}_{\leq t}$ denotes the history of {\textit{executed}} actions inferred from proprioceptive feedback, rather than previously commanded actions, accounting for imperfect low-level tracking.}, visual observations $\mathbf{I}_{\leq t}= \{\mathbf{I}_{t-m+1}, \dots, \mathbf{I}_t\}$, next actions $\mathbf{a}_{> t} = \{\mathbf{a}_{t+1}, \dots, \mathbf{a}_{t+n}\}$, where $n$ and $m$ represent the action and observation horizons, respectively. The \m{} framework transforms the distribution of historical actions directly into the distribution of future actions via a conditional flow in a shared latent space $\mathcal{Z}$.
	
	Fig. \ref{pipeline} details the \m{} architecture. Concretely, the proposed architecture leverages a Convolutional Neural Network (CNN)-based autoencoder to map action trajectories into a compact latent space $\mathcal{Z}$. Specifically, the action encoder $E_a$ and decoder $D_a$ parameterize the mapping $\mathbf{z}_0 = E_a(\mathbf{a}_{\leq t})$ and the reconstruction $\hat{\mathbf{a}}_{>t} = D_a(\mathbf{z}_1)$. Simultaneously, the visual encoder $E_I$ extracts features from multi-modal image streams $\mathbf{I}_{\leq t}$, which are further projected via an MLP to form a global conditioning vector $\mathbf{c} = {MLP}(E_I(\mathbf{I}_{\leq t}))$. Finally, we define the action-to-action flow in $\mathcal{Z}$ via a time-dependent vector field $\mathbf{v}_\tau$ that satisfies the ODE ${d\mathbf{z}_\tau}/{d\tau} = \mathbf{v}_\tau(\mathbf{z}_\tau)$ for $\tau \in [0, 1]$.
	
	Due to the physical consistency of sequential robot motions, adjacent action chunks exhibit inherent similarity. By further embedding these segments into a high-dimensional latent space and training with flow matching, the distribution of the starting point $\mathbf{z}_0$ is well aligned with the target $\mathbf{z}_1$ (see Figs. \ref{latent_space} and \ref{latent_space_appen}). This reduction in distributional distance drastically simplifies the transport mapping, enabling a lightweight MLP to attain strong performance even with single-step inference.
	
	Note that historical actions $\mathbf{a}_{\leq t}$ can also be corrupted with subtle noise to introduce stochasticity and preserve a certain degree of multimodality (Fig. \ref{multi_modal}). In the presence of action-level uncertainties, the performance of \m{} is compromised due to its inherent dependence on preceding action sequences. Injecting subtle stochastic perturbations into the historical actions before the encoding stage can substantially enhance its generalization capability against such uncertainties. Section \ref{sec-initial_state} provides empirical evidence supporting the effectiveness of this mechanism.

	\subsection{Learning objectives}
	
	The total training objective $\mathcal{L}_{total}$ is formulated as a multi-task loss to ensure generation accuracy and physical consistency simultaneously.
	
	\paragraph{Flow matching loss} The primary objective is the regression of the time-dependent vector field $f_\theta(\mathbf{z}_\tau, \tau, \mathbf{c})$ in the latent space $\mathcal{Z}$. This loss ensures that the model learns the optimal transport path between the latent starting action $\mathbf{z}_0$ and the latent target action $\mathbf{z}_1$, i.e.,
	\begin{equation}
		\begin{aligned}
			\mathcal{L}_{FM} = \mathbb{E}_{\tau \sim \mathcal{U}[0,1], \mathbf{z}_0, \mathbf{z}_1} \left\| f_\theta(\mathbf{z}_\tau, \tau, \mathbf{c}) - \mathbf{v}_\tau(\mathbf{z}_\tau, \tau, \mathbf{c}) \right\|^2.
			\label{action_flow_loss}
		\end{aligned}
	\end{equation}
	Eq. \eqref{action_flow_loss} is the specific \m{} formulation of Eq. \eqref{general_loss} in the action latent space $\mathcal{Z}$.
	
	\paragraph{Autoencoder reconstruction loss} To ensure that the latent space $\mathcal{Z}$ preserves the topological structure of the action space, we apply an $\ell_1$ reconstruction loss to the action autoencoder, i.e.,
	\begin{equation}
		\begin{aligned}
			\mathcal{L}_{AE} = \mathbb{E}_{\mathbf{a}_{> t}} \left\| \mathbf{a}_{> t} - D_a(E_a(\mathbf{a}_{> t})) \right\|_1.
		\end{aligned}
	\end{equation}
	This loss regularizes the encoder $E_a$ and decoder $D_a$ to maintain high-fidelity reconstruction of action chunks.
	
	\paragraph{Inference consistency loss} To bridge the gap between abstract latent generation and physical execution, inspired by~\citet{gao2025vita}, we introduce inference consistency loss. The inference consistency aims to align ODE-inferred and ground truth actions in both the latent space and the original action space, i.e.,
	\begin{equation}
		\begin{aligned} \label{loss_IC}
			\mathcal{L}_{IC} =  & \mathbb{E}_{\hat{\mathbf{z}}_1, \mathbf{a}_{> t}} \left\| \hat{\mathbf{z}}_1 - E_a(\mathbf{a}_{> t}) \right\|_1\\  & + \lambda_0\mathbb{E}_{\hat{\mathbf{z}}_1, \mathbf{a}_{> t}} \left\| D_a(\hat{\mathbf{z}}_1) - \mathbf{a}_{> t} \right\|_1 ,
		\end{aligned}
	\end{equation}
	where $\hat{\mathbf{z}}_1$ is the latent vector obtained via ODE integration and $\lambda_0\in \mathbb{R}_{>0}$ denotes a user-defined weight. This objective ensures that the generated flow trajectories translate into physically meaningful and executable actions.~\citet{gao2025vita} have found that $\mathcal{L}_{IC}$ is critical for avoiding latent space collapse.
	
	Finally, the total training objective $\mathcal{L}_{total}$ is formalized with three weighting coefficients $\lambda_1$, $\lambda_2$, and $\lambda_3 \in \mathbb{R}_{>0}$
	\begin{equation}
		\begin{aligned}
			\mathcal{L}_{total} = \lambda_1\mathcal{L}_{FM} + \lambda_2\mathcal{L}_{AE} + \lambda_3\mathcal{L}_{IC}.
		\end{aligned}
	\end{equation}
	
	\section{Evaluation}
	
	\begin{table*}[ht]
		\caption{Success rates across 5 simulation tasks under 9 different algorithms (100 demonstrations, 30 epochs). The best results are highlighted in {bold}, while the second-best results are indicated with {underlines}.}
		\label{tab-30epochs}
		\centering
		\small
		\setlength{\tabcolsep}{3.5pt} 
		\begin{tabular}{lcccccc}
			\toprule
			\textbf{Methods} & \textbf{Steps} & \textbf{Close Box} & \textbf{Pick Cube} & \textbf{Stack Cube} & \textbf{Open Drawer} & \textbf{Pick-Place Bowl} \\
			& & (\%) & (\%) & (\%) & (\%) & (\%) \\
			\midrule
			\textbf{\m{}} & 6 & \textbf{92} & \textbf{92} & \textbf{86} & \textbf{92} & \underline{90} \\
			VITA                & 6 & \underline{88} & \underline{88} & \underline{80} & \underline{90} & \textbf{92} \\
			FM-UNet             & 10 & 82 & 70 & 28 & 34 & 68 \\
			FM-DiT              & 10 & 58 & \underline{88} & 26 & 28 & 84 \\
			DDPM-UNet           & 100 & 72 & 60 & 36 & 64 & 66 \\
			DDPM-DiT            & 100 & 58 & 58 & 16 & 14 & 68 \\
			DDIM-UNet           & 40 & 70 & 56 & 36 & 64 & 82 \\
			Score-UNet          & 100 & 36 & 36 & 12 & 0 & 4 \\
			ACT                 & 1 & 82 & 86 & 32 & 80 & 60 \\
			\bottomrule
		\end{tabular}
	\end{table*}
	
\begin{figure*}
		\centering
		\includegraphics[width=1\linewidth]{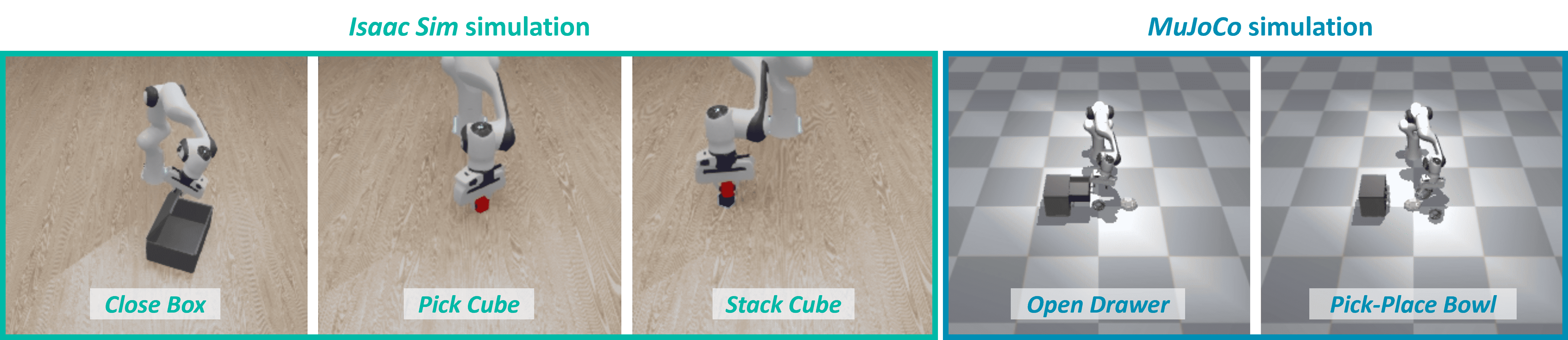}
		\caption{\textbf{Simulational tasks.} Simulations are conducted in the \textit{Roboverse} platform~\citep{geng2025roboverse}. Implementary tasks include \textit{Stack Cube} and \textit{Pick Cube} from ManiSkill~\citep{mu2021maniskill}, \textit{Close Box} from RLBench~\citep{james2020rlbench}, \textit{Open Drawer} and \textit{Pick-Place Bowl} from LIBERO~\citep{liu2023libero}).}
		\label{experiments}
	\end{figure*}

	We conduct a comprehensive evaluation of the proposed \m{} across 5 simulational tasks (\textit{Stack Cube} and \textit{Pick Cube} from ManiSkill~\citep{mu2021maniskill}, \textit{Close Box} from RLBench~\citep{james2020rlbench}, \textit{Open Drawer} and \textit{Pick-Place Bowl} from LIBERO~\citep{liu2023libero}) in \textit{Roboverse} platform~\citep{geng2025roboverse} and 2 real-world tasks (\textit{Pick Cube} and \textit{Open Drawer}) on \textit{Franka} robot, as shown in Fig. \ref{experiments}, Fig. \ref{real_test}, and Fig. \ref{real_test_close}. Our performance is benchmarked against eight state-of-the-art baseline methods, including DDPM-UNet~\citep{dp, ddpm}, DDPM-DiT~\citep{ddpm, Peebles_2023_ICCV}, DDIM-UNet~\citep{dp, song2020denoising}, FM-UNet~\citep{flowmatching}, FM-DiT~\citep{flowmatching, Peebles_2023_ICCV}, Score-UNet~\citep{song2020score}, ACT~\citep{zhao2023learning}, and VITA \citep{gao2025vita}. To ensure a fair and rigorous comparison, we have standardized the hyperparameters (e.g., chunk size, batch size) and network scales across all evaluated methods to the greatest extent possible. The evaluation primarily focuses on training efficiency, inference cost, and generalization. We use $\lambda_0 = 0.5$, $\lambda_1 = 1$, $\lambda_2 = 0.5$, $\lambda_3 = 1$, $n = m = 8$, and a batch size of 32 across all experiments (Appendix~\ref{appen_param}).
	
	\subsection{Training efficiency}
	
	\begin{wrapfigure}{r}{0.55\textwidth}
		\centering
		\includegraphics[width=1\linewidth]{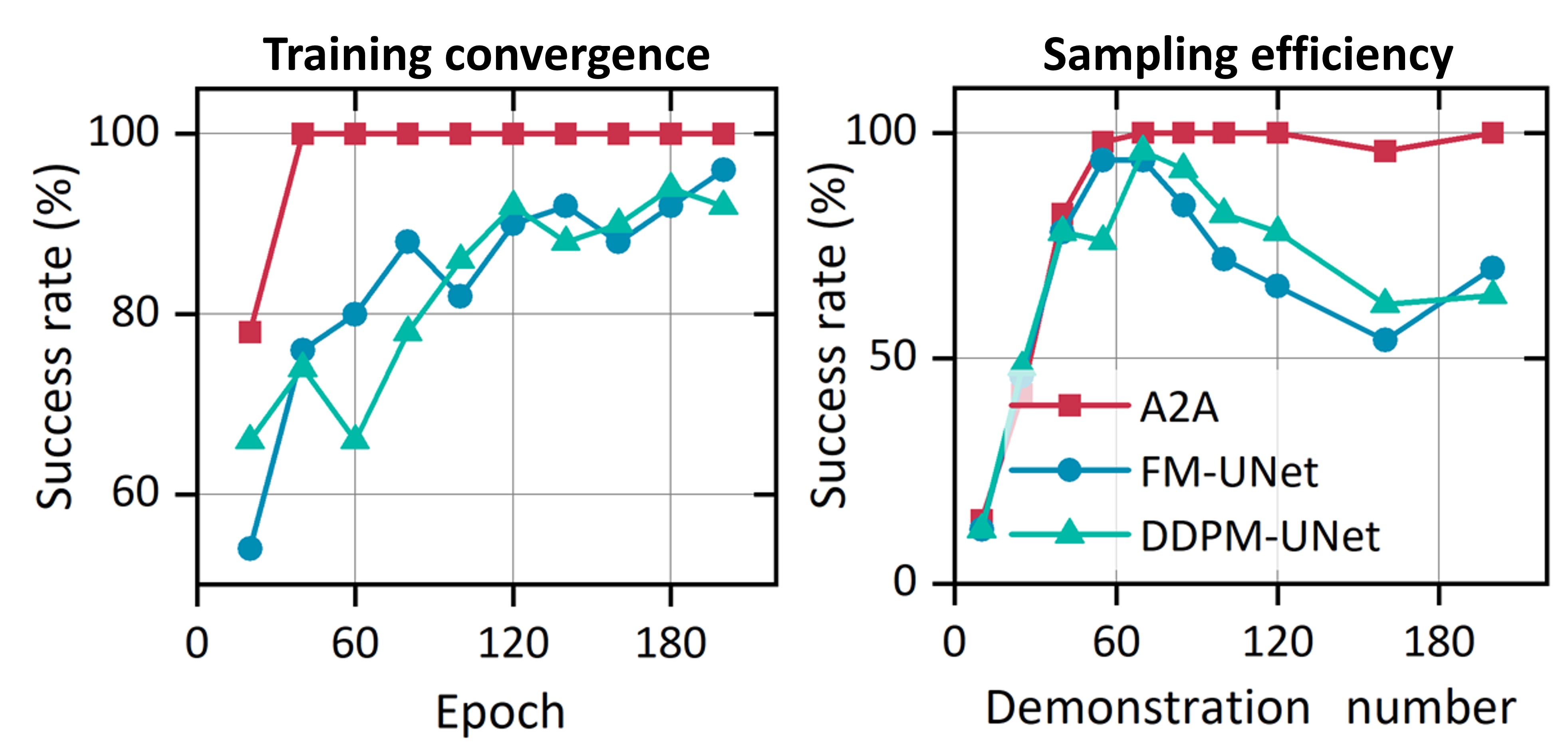}
		\caption{\textbf{Training efficiency test.} \textbf{Left:} Success rates across varying training epochs (using 100 demonstrations in \textit{Close Box} task). \textbf{Right:} Success rates across varying demonstration numbers (100 epochs in \textit{Stack Cube} task).}
		\label{training_efficiency}
	\end{wrapfigure}
	
	We first analyze the performance of \m{} across different training data sizes and epoch numbers. Fig. \ref{training_efficiency} illustrates the training efficiency of all evaluated methods. As illustrated in Fig. \ref{training_efficiency} (Left), \m{} demonstrates superior convergence speed compared to DDPM-UNet and FM-UNet, achieving a stable 100\% success rate within significantly limited 40 training epochs on the \textit{Close Box} task. 
	
	Furthermore, the sampling efficiency results in Fig. \ref{training_efficiency} (Right) reveal that \m{} quickly reaches and maintains a high performance ceiling as the number of demonstrations increases. In contrast, both DDPM-UNet and FM-UNet exhibit noticeable fluctuations and lower stability. This discrepancy likely stems from the increasing trajectory diversity in the increasing dataset of the \textit{Stack Cube} task, which may require higher model capacity or extended training epochs to accommodate. To validate this hypothesis, we conduct further evaluations on the DDPM-UNet and FM-UNet baselines. As illustrated in Fig. \ref{iterations_ddpm_fm}, their success rates gradually converge to 100\% as training epochs increase. 
	
	\begin{figure*}
		\centering
		\includegraphics[width=1\linewidth]{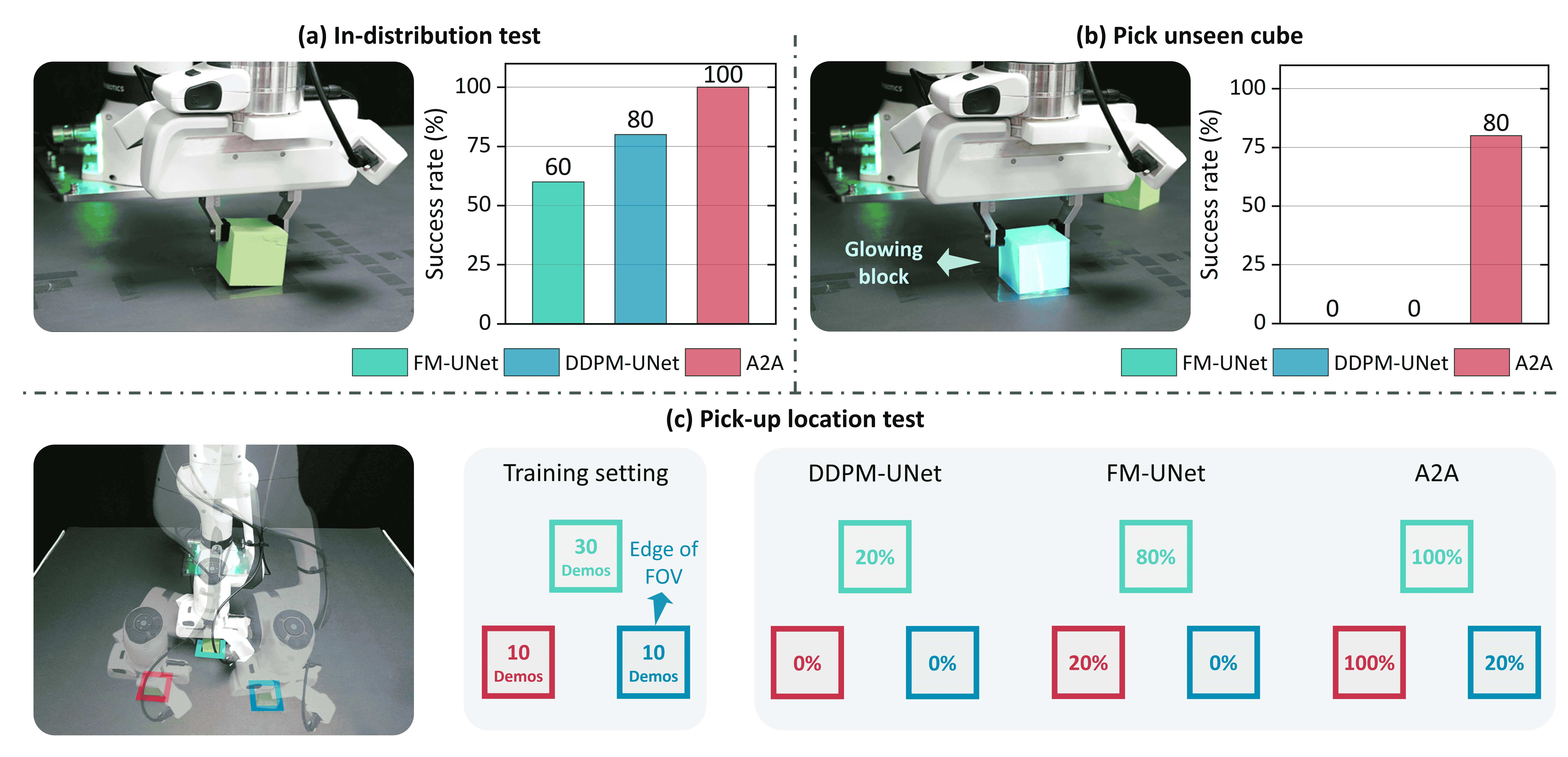}
		\caption{\textbf{Experimental results of \textit{Pick Cube} task.} (a) Policies are trained on a limited dataset of 30 trajectories for 100 epochs. During evaluation, each method is tested over 10 trials. (b) Generalization capability is further challenged by replacing the target with an unseen glowing block. (c) Pick the cube from different locations with a limited 10 training demonstrations.}
		\label{real_test}
	\end{figure*}
	
	\begin{wrapfigure}{r}{0.55\textwidth}
		\centering
		\includegraphics[width=1\linewidth]{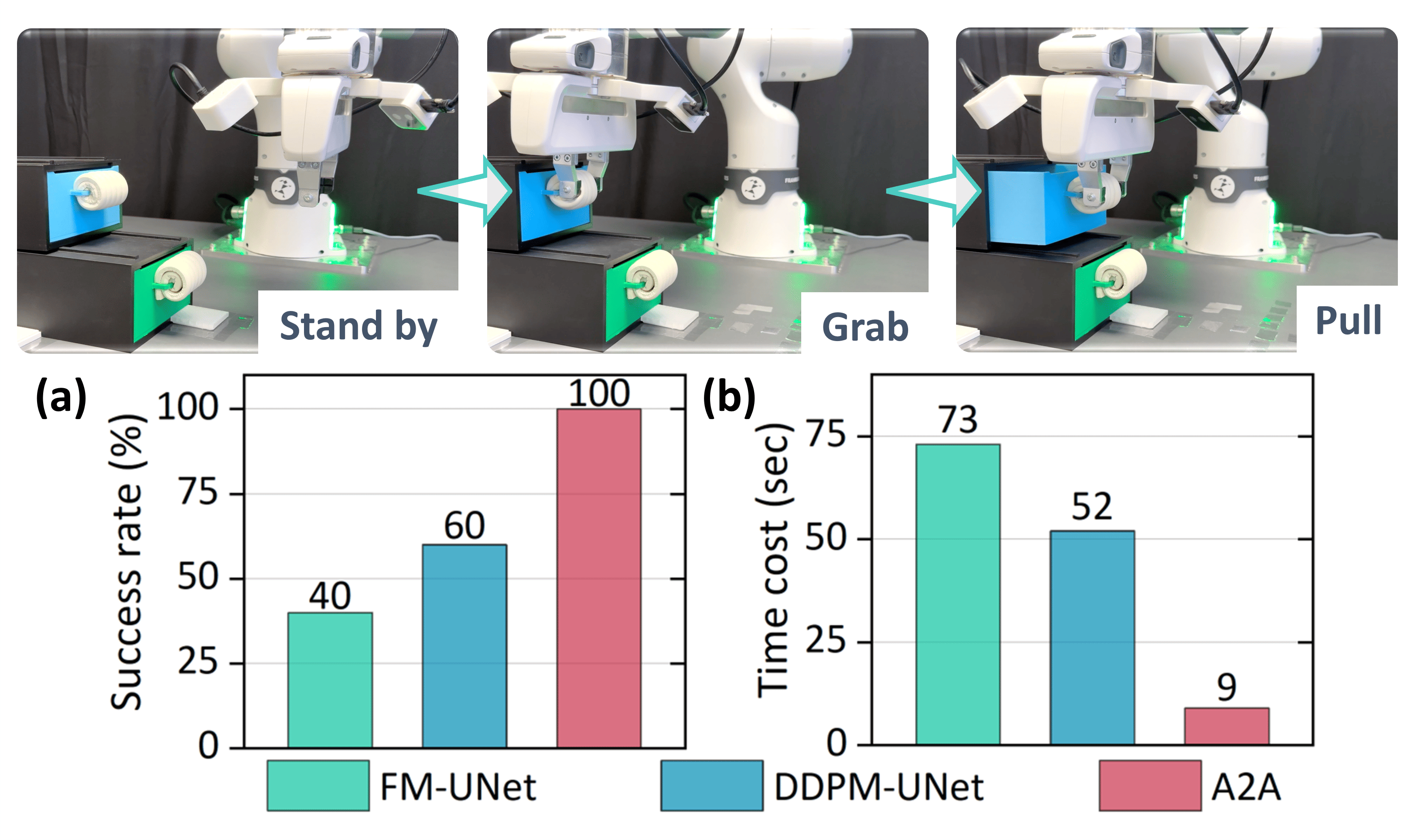}
		\caption{\textbf{Experimental results of \textit{Open Drawer} task.} Policies are trained on a limited dataset of 30 trajectories for 300 epochs. Time cost denotes the total time elapsed during task completion. Success rate is evaluated over 10 trials.}
		\label{real_test_close}
	\end{wrapfigure}
	
	Real-world test in Fig. \ref{real_test} (a) and Fig. \ref{real_test_close} (a) further shows that with only 30 training trajectories, A2A-Flow achieves a 100\% in-distribution success rate, outperforming DDPM-UNet and FM-UNet. Fig. \ref{real_test_close} (b) further illustrates that in the \textit{Open Drawer} task, the proposed algorithm achieves a shorter completion time, whereas DDPM-UNet and FM-UNet exhibit significant hesitation during the operation. Moreover, we further implement a pick-up location test. Two additional pick-up locations are added (Fig. \ref{real_test} (c)), each with only 10 demonstrations. \m{} demonstrates rapid adaptation and high success rates, showcasing its superior data efficiency in novel scenarios. Even at the edge of the field-of-view (FOV) (bottom-right location, Fig. \ref{camera_view}), \m{} sustains a viable success rate, underscoring its robustness to suboptimal visual inputs. 
	These results indicate that \m{} policy achieves superior performance, particularly in regimes characterized by limited training data and fewer training epochs.

	The quantitative results across 5 different tasks and 9 algorithms are summarized in Table~\ref{tab-30epochs}. \m{} consistently achieves the highest success rates. VITA, which similarly incorporates an inference consistency mechanism~\citep{gao2025vita}, also achieves a competitively high success rate. Notably, we find that given advanced transformer architectures and identical hyperparameters, regression-based ACT achieves performance comparable to generation-based methods. This observation aligns with recent findings by~\citet{pan2025much}.
	
	\subsection{Inference cost}
\begin{figure*}
		\centering
		\includegraphics[width=0.9\linewidth]{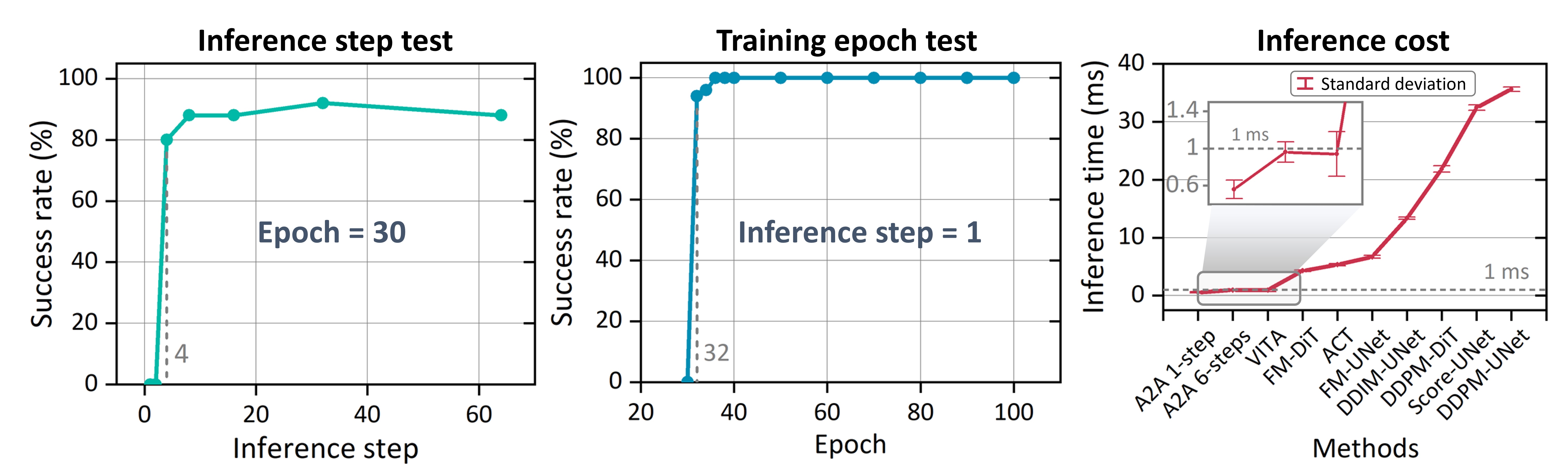}
		\caption{\textbf{Inference cost test.} Setting: \textit{Close Box}. \textbf{Left:} Success rate over inference steps with fixed 30 epochs. \textbf{Middle:} Success rate over training epochs with only one-step inference. \textbf{Right:} Mean inference time per sampling step for all evaluated models, benchmarked on an identical hardware to ensure fairness.}
		\label{inference_cost}
	\end{figure*}

	Capitalizing on the inherent efficiency of flow matching, we further examine the extreme inference speed achievable by the \m{} policy. We first evaluate the impact of sampling steps on training performance. As shown in Fig. \ref{inference_cost} (Left), increasing the number of inference steps leads to a rapid improvement in success rates; however, the marginal gains diminish significantly beyond 4 steps. Regarding Fig. \ref{inference_cost} (Middle), when the inference budget is restricted to only one step, the success rate rises above 90\% substantially after 32 training epochs.
	
\begin{figure*}
		\centering
		\includegraphics[width=0.9\linewidth]{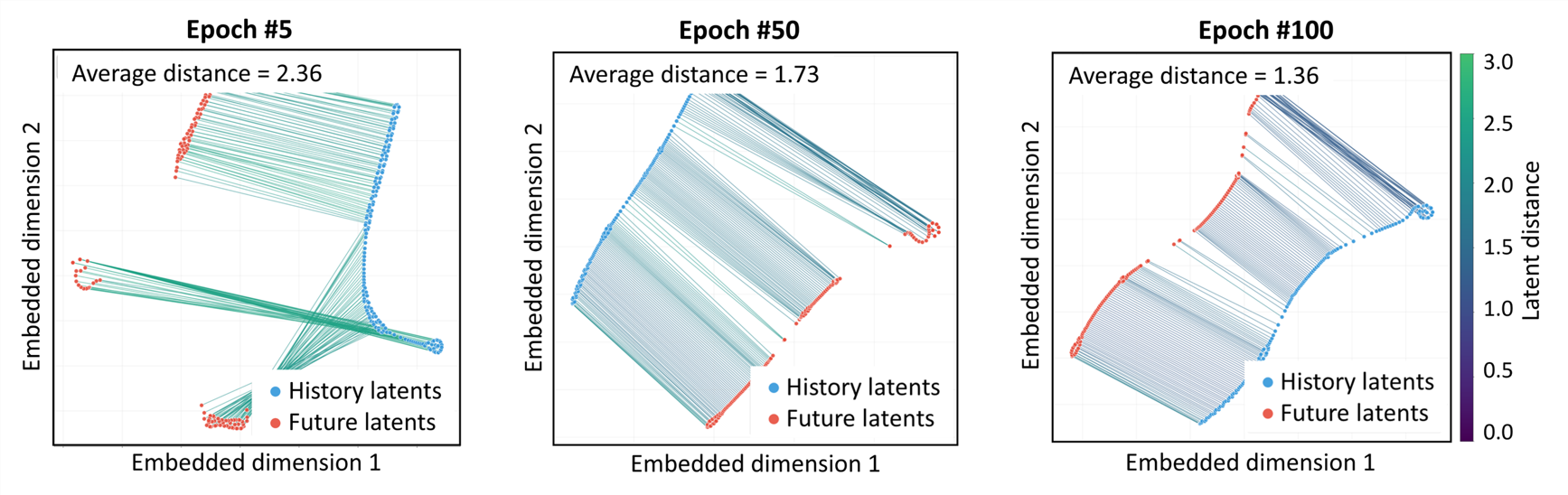}
		\caption{\textbf{Convergence of latent space representations during \m{} training.} Settings: \textit{Close Box}, one-step inference. We apply t-SNE to jointly embed history and future action latents, with paired samples connected by lines. Line colors indicate distances computed in the 512-dimensional latent space.}
		\label{latent_space}
	\end{figure*}

	Fig. \ref{latent_space} visualizes the convergence of latent space representations under the one-step inference during training. We employ t-SNE to jointly embed history and future action latents, with paired samples connected by lines to represent the learned flow. As training progresses, the average distance between the history and future action chunks decreased significantly. Furthermore, the trajectories connecting these pairs increasingly align into parallel paths. These phenomena provide strong empirical evidence for the feasibility of single-step flow mapping from history to future latents, while the emerging parallelism underscores the rectilinearity of the learned flow.
	
	Furthermore, we benchmark the mean inference time per sampling step across various algorithms on identical hardware (NVIDIA GeForce RTX 5090 GPU with 32GB VRAM), as depicted in Fig. \ref{inference_cost} (Right). Attributable to the extreme compressibility of sampling steps and the efficient MLP-based architecture, the inference latency of \m{} is maintained below 1 ms. Notably, in the single-step inference regime, the latency reaches an impressive 0.56 ms, indicating significant potential for tasks that demand high-frequency decision-making.
	
	Note that \m{} and VITA~\citep{gao2025vita} exhibit superior inference speed compared to regression-based ACT~\citep{zhao2023learning}. Beyond the reduced number of sampling epochs, this efficiency also stems from the pure MLP operations in the latent space. Conversely, ACT in this work utilizes a cost Transformer architecture with self-attention and cross-attention mechanisms.
	
	
	\begin{table}[ht]
		\caption{Comparison of success rates across 4 different randomization scenes. All models are trained with 100 demonstrations for 200 epochs in Level 0. The best results are highlighted in {bold}, while the second-best results are indicated with {underlines}.}
		\label{tab-generalization}
		\centering
		\small
		\setlength{\tabcolsep}{3.5pt}
		\begin{tabular}{lcccc}
			\toprule
			\textbf{Methods} & \textbf{Level 0} & \textbf{Level 1} & \textbf{Level 2} & \textbf{Level 3} \\
			& (\%) & (\%) & (\%) & (\%) \\
			\midrule
			\textbf{\m{} (1 step)} & \textbf{100} & \underline{20} & \underline{16} & \underline{22} \\
			\textbf{\m{} (6 steps)} & \textbf{100}  & \textbf{38} & \textbf{42} & \textbf{38} \\
			VITA                & \textbf{100}    & 4           & 2           & 2          \\
			FM-UNet             & \underline{96}  & 6           & 6           & 4          \\
			DDPM-UNet           & 92              & 2          & 4           & 2          \\
			Score-UNet          & 94              & 0          & 2           & 0          \\
			ACT                 & 86              & 8          & 2           & 0          \\
			\bottomrule
		\end{tabular}
	\end{table}
	
	\subsection{Generalization performance}
	
	\subsubsection{Visual uncertainty}
	
	We further evaluate the generalization performance of the proposed \m{} policy under various visual uncertainties. \textit{Roboverse}~\citep{geng2025roboverse}, the adopted platform, categorizes the scene randomizations for the \textit{Close Box} task into four progressive levels of difficulty. Level 0 serves as the training dataset, involving initial box pose variations (see Appendix Fig. \ref{level0}). Level 1 introduces significant background textures randomization (see Appendix Fig. \ref{level1}), while Level 2 adds illumination perturbations. Finally, Level 3 incorporates camera viewpoint variations. Detailed randomization configurations are provided in the Appendix \ref{appen_dr}.
	
	\begin{wrapfigure}{r}{0.55\textwidth}
		\centering
		\includegraphics[width=1\linewidth]{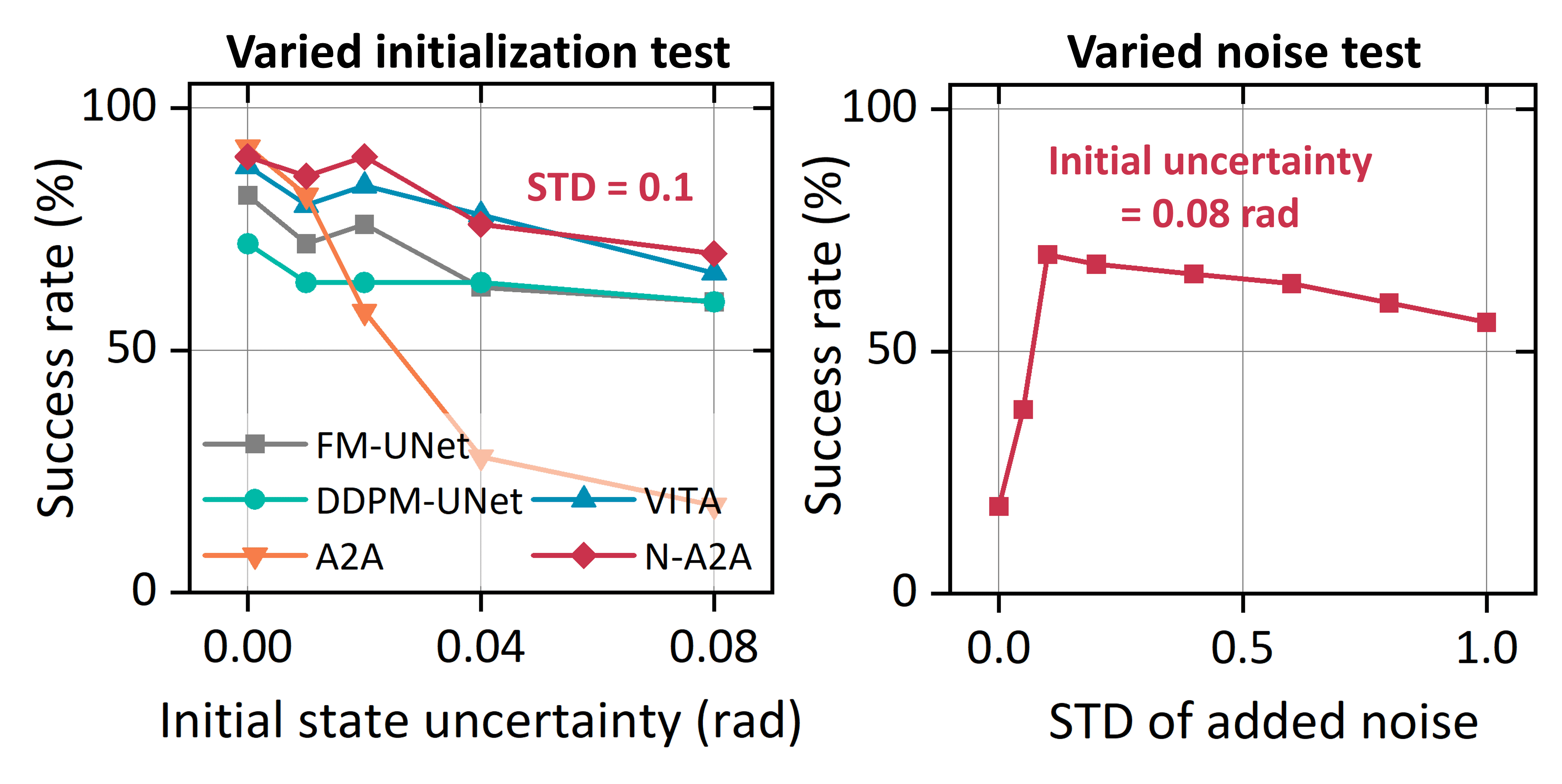}
		\caption{\textbf{Generalization test on different initiation.} \textbf{Left:} Success rate under varying levels of initial state uncertainty. Settings: \textit{Close Box}, 30 epochs, 100 demonstration numbers. Noised \m{} (N-\m{}) refers to the initial action distribution occupied by 0.1 {STD} \textit{Gaussian} noise. \textbf{Right:} Success rate under varying levels of initial noise. The initial state uncertainty is set as 0.08 rad.}
		\label{FIGinitial}
	\end{wrapfigure}
	
	Table \ref{tab-generalization} presents the success rates of evaluated methods across Levels 0 to 3. Notably, even when encountering Level 1-3 for the first time, \m{} (6 steps) maintains a robust success rate of 30-40\%, consistently outperforming all baseline methods. In the single-step inference regime, \m{} continues to exhibit superior generalization compared to other algorithms. We also verified its visual generalization performance in real-world tests, as shown in Fig. \ref{real_test}. We substitute the targeted cube with an unseen glowing variant, inducing severe visual distractors. The baselines fail entirely in this case, whereas our algorithm sustains a robust 80\% success rate.
	
	We argue that the fundamental reason for this robustness is our decoupled strategy, lightening the \textit{representation entanglement} problem \citep{li2026causal}. Unlike conventional methods that simply concatenate proprioceptive and visual features, we process them through distinct strategies, preventing low-dimensional proprioceptive signals from being overshadowed by high-dimensional ones, thereby enabling the model to leverage the complementary strengths of each modality more effectively. Specifically, grounding the generation process in historical actions can substantially improve robustness to visual perturbations, which enforces physical consistency over time. Moreover, we find that injecting noise into historical trajectories (0.02 standard deviation, STD) can further narrow the generalization gap, from $20\%$ to $52\%$ on Level 1.
	
	\subsubsection{Initial state uncertainty} \label{sec-initial_state}
	
	Given the temporal dependency of subsequent action sequences on previous ones in \m{}, a natural question arises regarding its robustness to uncertainties in the historical sequence. To investigate this, we randomize the initial pose of the robot, as illustrated in Fig. \ref{initial_state}. The results in Fig. \ref{FIGinitial} (Left) indicate that the \m{}, compared to baselines, which generate actions from pure noise at each step, is indeed more sensitive to uncertainties within the action history. However, by injecting a small amount of \textit{Gaussian} noise (0.1 standard deviation, {STD}) into the historical actions, it can be observed that a significant boost in generalization performance is achieved. Fig. \ref{FIGinitial} (Right) further depicts the relationship between the success rate and the intensity of the injected noise. How to optimally fuse clean historical data with \textit{Gaussian} noise to balance determinism and stochasticity remains a compelling direction for future research.
	
	\section{Ablation study}
	We further conduct ablation studies on the architectural design to answer two fundamental questions: whether the generative paradigm provides superior performance over the deterministic regression baseline; whether performing flow matching within the latent space is more effective than directly in the raw action space.
	
	\subsection{Regression or generation}
	
	\begin{wrapfigure}{r}{0.6\textwidth}
		\centering
		\includegraphics[width=1\linewidth]{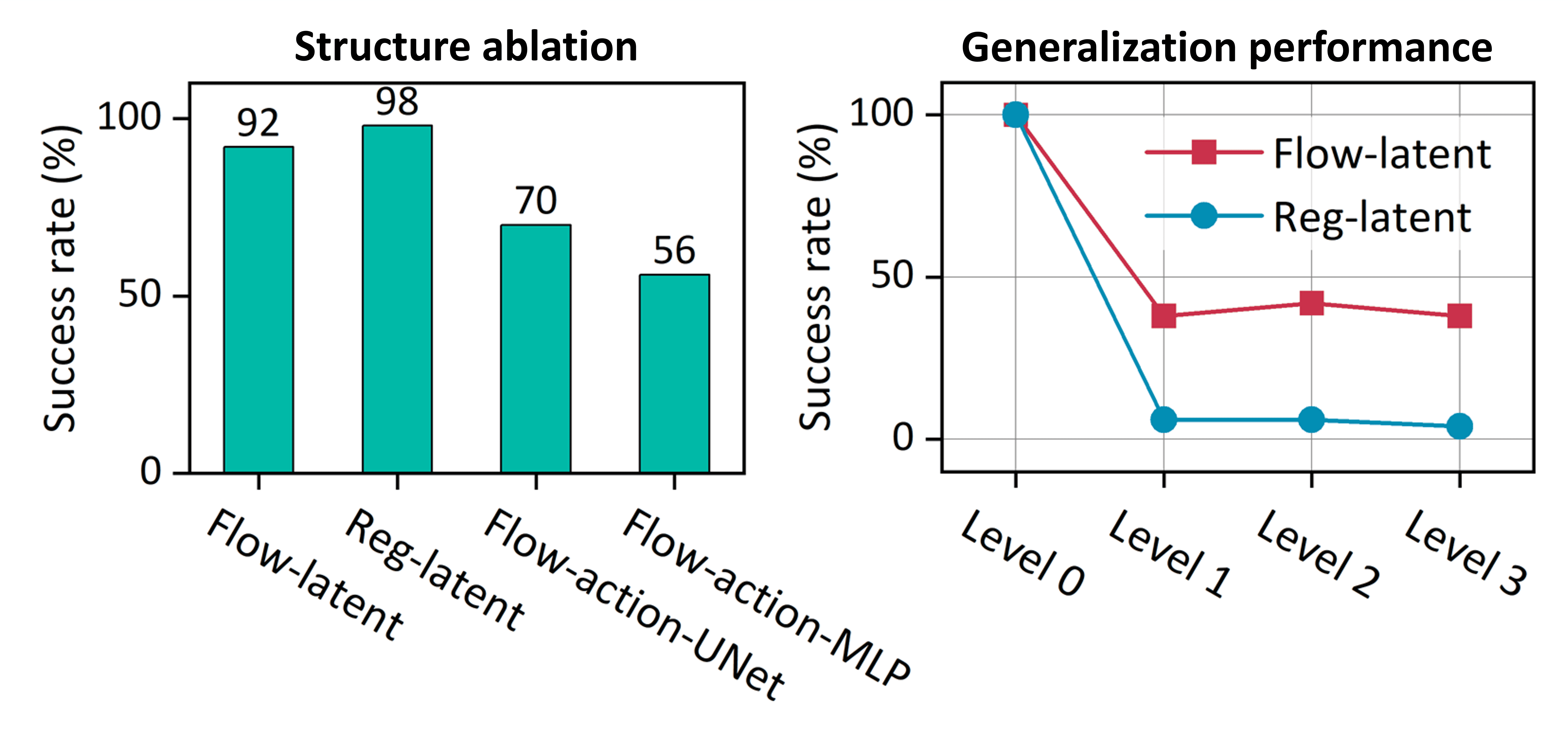}
		\caption{\textbf{Ablation study of model structure.} Settings: \textit{Close Box}, 30 epochs, 100 demonstration numbers, 6 inference steps for flow-based methods. \textbf{Left:} Impact of learning objectives and representation spaces. Comparison of flow matching and regression strategies implemented in both latent and raw action spaces. \textbf{Right:} Generalization capability. Robustness comparison between latent-space regression and flow matching under varying environmental perturbations.}
		\label{structure_ablation}
	\end{wrapfigure}
	
	There is a growing discourse regarding the relative merits of \textit{generation} versus \textit{regression} in the context of robotic control~\citep{pan2025much}. Here, we also attempt to substitute the flow matching objective with a deterministic regression approach. To ensure a fair comparison, all other architectural components, including the encoder and the latent space configuration, remain strictly identical. The results are presented in Fig. \ref{structure_ablation} (Left), where \textit{Flow-latent} denotes the flow matching performed within the latent space, i.e., our final choice. \textit{Reg-latent} represents deterministic regression performed within the latent space. We found that both methods achieve high success rates on the training distribution, which aligns with recent findings by~\citet{pan2025much}. However, Fig. \ref{structure_ablation} (Right) reveals that the generative approach exhibits significantly higher resilience to environmental perturbations, whereas the regression variant fails to generalize to unseen scenarios. 
	This gap might stem from the decoupling of action and visual inputs. Direct combination with higher-dimensional visual representations for the regression method may dilute the benefits of low-dimensional proprioceptive signals.  
	
	\subsection{Action space or latent space}
	We further evaluate flow matching performed without the latent space, using both U-Net and MLP (same as \m{}) backbones. The results are shown in Fig. \ref{structure_ablation} (Left), where \textit{Flow-action-UNet} represents flow matching directly in the raw action space using a U-Net architecture, and \textit{Flow-action-MLP} denotes flow matching directly in the raw action space using an MLP architecture. It can be seen that flow matching in the raw action space leads to inferior convergence performance compared to the latent-space approach. This reason can be attributed to the high-dimensional representation in the latent space, which effectively aligns the initial and target distributions of the flow. This structured alignment facilitates a smoother learning process, enabling the model to achieve high performance even within a single-step inference regime, as shown in Figs. \ref{latent_space} and \ref{latent_space_appen}.
	
	\begin{figure}
		\centering
		\includegraphics[width=0.5\linewidth]{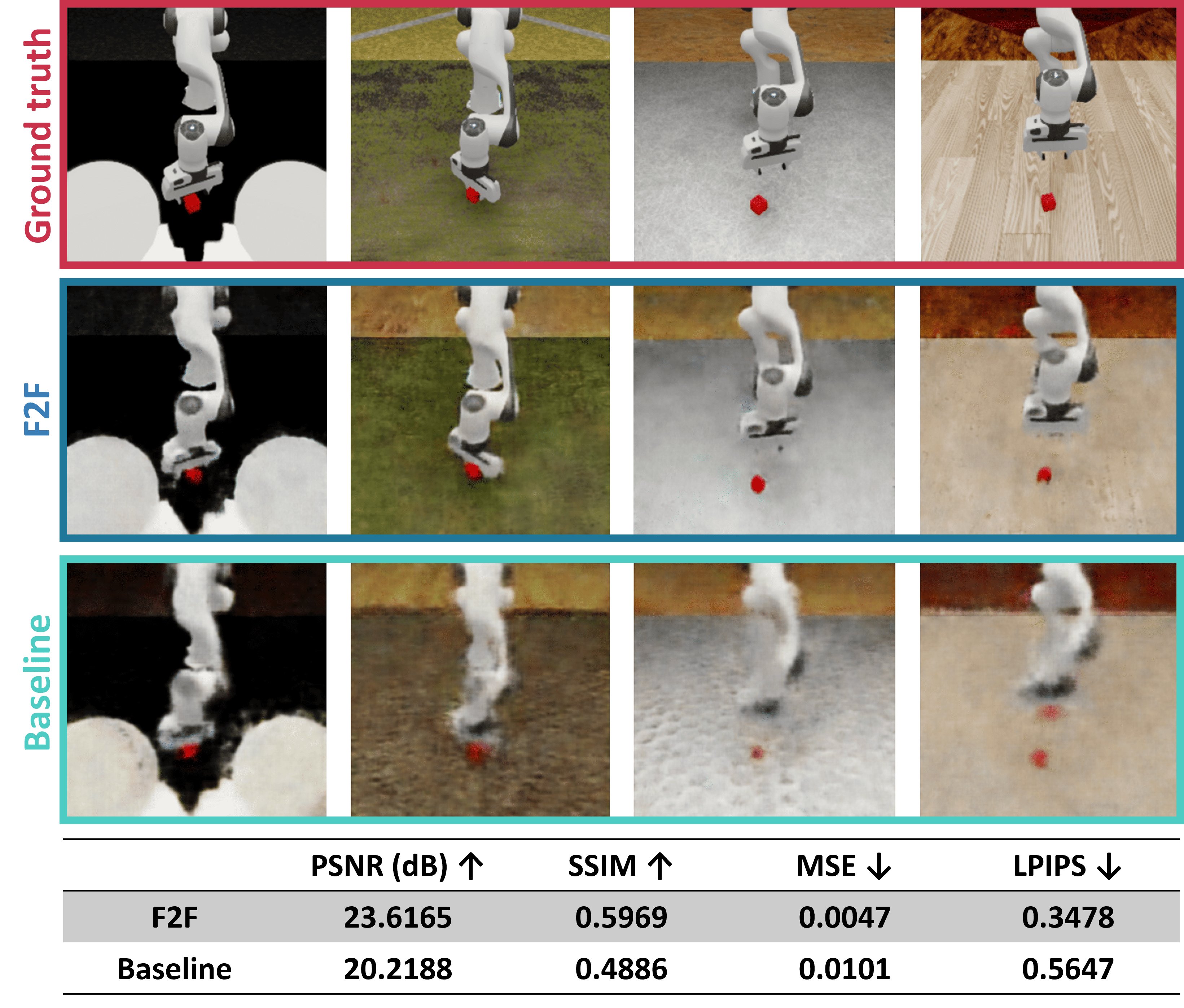}
		\caption{\textbf{Video generation results.} The predicted third frames in four different unseen scenarios are visualized.}
		\label{video_generation}
	\end{figure}
	
	\section{Application to video generation}
	
	Similar to robotic manipulation, video generation in the form of future frame prediction is inherently a temporal continuity task. High-fidelity future video prediction can enhance the performance of VLA models~\citep{zhao2025cot, deng2026video}. This work further explores the transferability of the A2A paradigm to video generation, hereafter referred to as \textbf{F}rames-to-\textbf{F}rames flowing matching (\textbf{F2F}). The training dataset comprises 100 videos for each level (Levels 0–4) of the \textit{pick cube} task, while the test set consists of four unseen scenarios in the same task. F2F is designed to predict three future frames based on a history of three consecutive frames. Both F2F and baseline are trained with 500 epochs. See Appendix \ref{appen_video} for implementation details. As illustrated in Fig. \ref{video_generation}, F2F achieves significantly higher generation quality compared to a regression-based baseline implemented under the same network configuration. While these results are achieved with a small-scale model, the F2F paradigm possesses significant scaling potential for larger architectures. Future work will further pursue the integration of predicted video into the policy architecture to further augment the performance of \m{}.
	
	\input{sec/conclusion}
	
	\clearpage
	\newpage
	\bibliographystyle{assets/plainnat}
	\bibliography{paper}
	
	\clearpage
	\newpage
	
	\appendix
	
	\section{Appendix}
	
	\subsection{Randomization level setting} \label{appen_dr}
	
	To systematically evaluate the robustness of evaluated methods, we utilize \textit{Roboverse}'s hierarchical generalization levels by introducing stochastic perturbations to the simulation environment. This section details the configuration of Level 0 (object randomization), Level 1 (+background randomization), Level 2 (+lighting randomization), and Level 3 (+camera viewpoint randomization).
	
	Level 0 introduces the randomization of initial object positions, as illustrated in Fig. \ref{level0}. It serves as the training set for the evaluated algorithms.
	
	Level 1 primarily focuses on the randomization of the environmental background, as illustrated in Fig. \ref{level1}. This level is designed to evaluate the policy's robustness against non-task-relevant visual distractors.
	
	Level 2 further adds lighting randomization. We randomize illumination using primary DiskLights (12k-45k intensity), auxiliary SphereLights (5k-20k intensity), and ambient presets. Parameters include color temperatures from 2500K to 6500K, directional jitters of $\pm 15^\circ$ for ceiling lights, and sphere light positional offsets of $\pm 0.5$m (lateral/longitudinal) and $\pm 0.3$ m (vertical) to create diverse shadowing patterns.
	
	Level 3 implements camera viewpoint randomization additionally. Camera extrinsics are perturbed using a uniform distribution within a delta range of $\pm 20$cm for lateral/longitudinal shifts. Vertical shifts are restricted to an upward range of $0$ to $10$ cm.

	\subsection{Hyperparameters} \label{appen_param}
	
	Training hyperparameters used in all simulations and experiments are set to the same, as shown in Table \ref{tab-parameters}. Note that the standard ACT implementation on the \textit{Roboverse} platform comprises approximately 60M parameters, nearly double that of DDPM-UNet ($\approx$ 28M). To ensure a fair comparison, we have modified the Transformer backbone of ACT to halve its parameter count, thereby aligning the model scales across all evaluated baselines.
	
	\renewcommand*{\thetable}{S1}
	\begin{table}[ht]
		\caption{Training hyperparameters.}
		\label{tab-parameters}
		\centering
		\begin{tabular}{lc}
			\toprule
			\textbf{Hyperparameters} & \textbf{Value} \\
			\midrule
			n & 8 \\
			m & 8  \\
			$\lambda_0$          & 0.5 \\
			$\lambda_1$          & 1   \\
			$\lambda_2$          & 0.5  \\
			$\lambda_3$          & 1   \\
			Batch size           & 32   \\
			\bottomrule
		\end{tabular}
	\end{table}
	
	\subsection{Experimental setup}
	In real tests, we deploy \m{} on a Franka robotic platform, maintaining complete consistency in training parameters with our simulation baseline. A key distinction from the simulation environment is the utilization of dual-view visual input, the setup of which is shown in Fig. \ref{camera_view}.
	
	\renewcommand*{\thefigure}{S1}
	\begin{figure}
		\centering
		\includegraphics[width=0.45\linewidth]{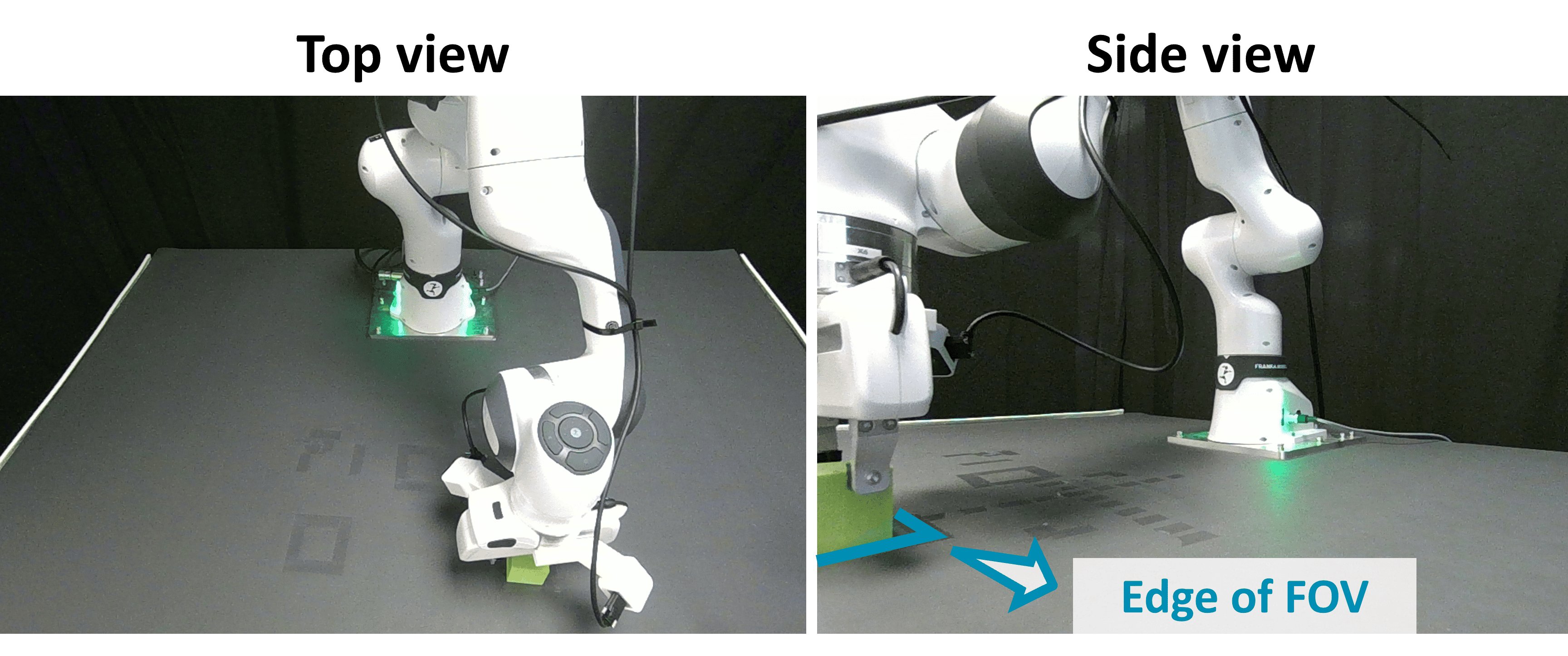}
		\caption{\textbf{The camera views in real tests.}}
		\label{camera_view}
	\end{figure}
	
	\subsection{Video generation} \label{appen_video}
	Fig. \ref{pipeline_f2f} illustrates the architectural framework of the F2F algorithm, showcasing the transition from historical frame sequences to future predictions. Historical frames $I_{\leq t}$ are encoded into a 512-dimensional latent space using a ResNet18 backbone and a VAE head to obtain the initial state $z_0$. A Flow Net Transformer, consisting of 4 layers and 4 attention heads, learns the vector field $v$ to map $z_0$ to the target latent $z_1$. The future frames $I_{> t}$ are then reconstructed through a 5-layer convolutional upsampling decoder. We employ a deterministic regression model as the baseline, which predicts future frames directly from historical sequences. For a fair comparison, aside from the omission of the flow-matching training objective, the baseline maintains an architecture identical to F2F.
	
	During evaluation, the predictive performance is assessed using four complementary metrics: Peak Signal-to-Noise Ratio (PSNR), Structural Similarity Index Measure (SSIM), Mean Squared Error (MSE), and Learned Perceptual Image Patch Similarity (LPIPS).
	
	\renewcommand*{\thefigure}{S2}
	\begin{figure}
		\centering
		\includegraphics[width=0.5\linewidth]{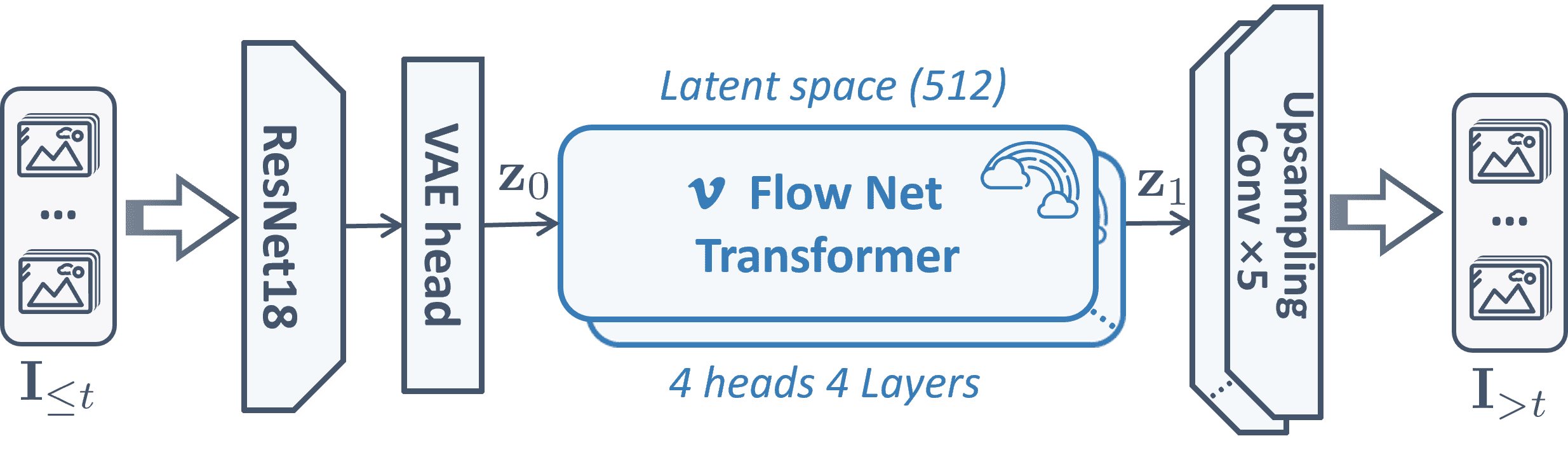}
		\caption{\textbf{Overview of F2F architecture for video prediction.} The model leverages a ResNet-VAE architecture to compress historical observations into a latent space, where a Transformer-based flowing matching computes the transport to future states. The predicted sequence $I_{> t}$ is generated via a convolutional upsampling block.}
		\label{pipeline_f2f}
	\end{figure}
	
	\renewcommand*{\thefigure}{S3}
	\begin{figure*}
		\centering
		\includegraphics[width=0.35\linewidth]{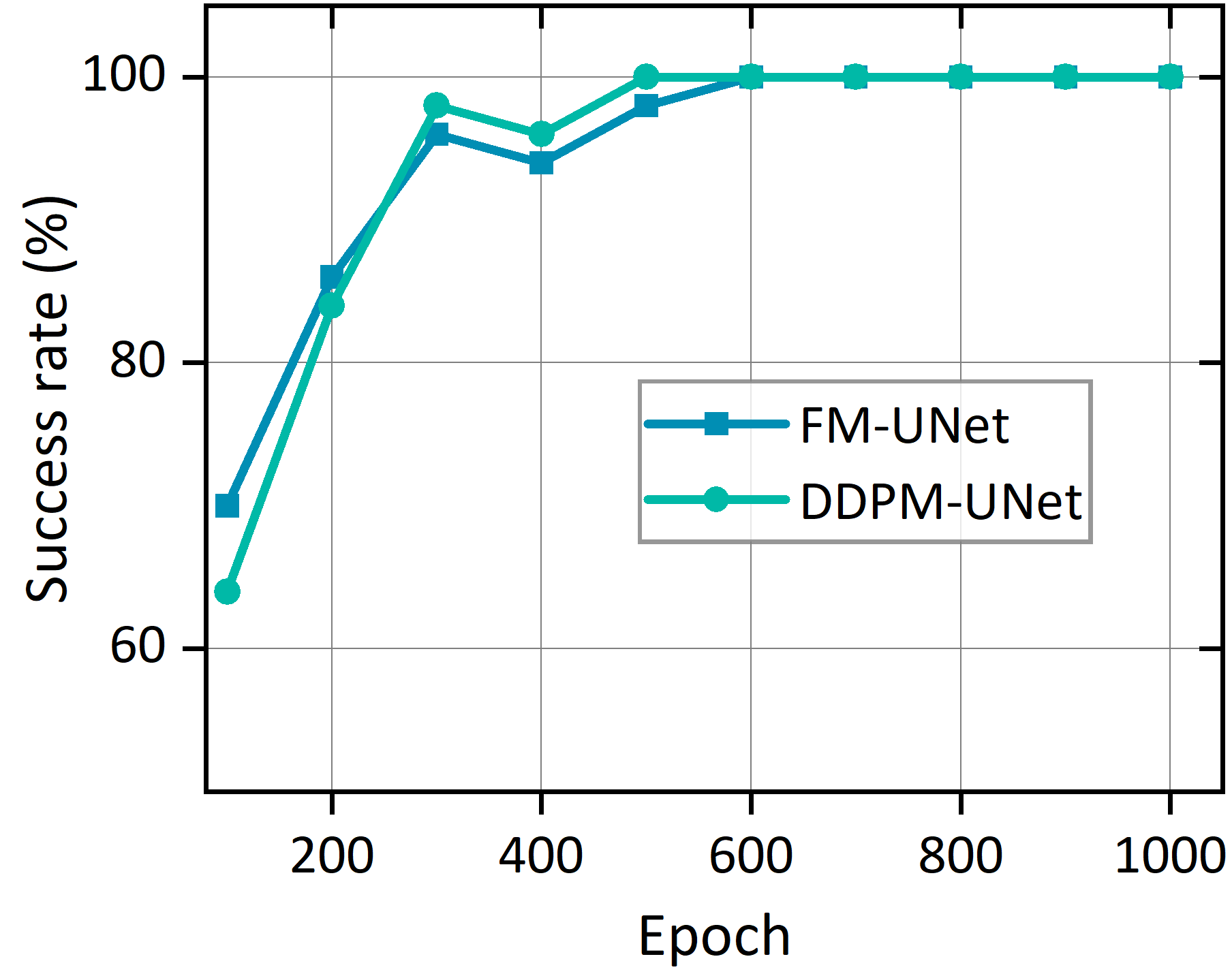}
		\caption{\textbf{Additional test on training efficiency of DDPM-UNet and FM-UNet.}}
		\label{iterations_ddpm_fm}
	\end{figure*}
	
	\renewcommand*{\thefigure}{S4}
	\begin{figure*}
		\centering
		\includegraphics[width=0.7\linewidth]{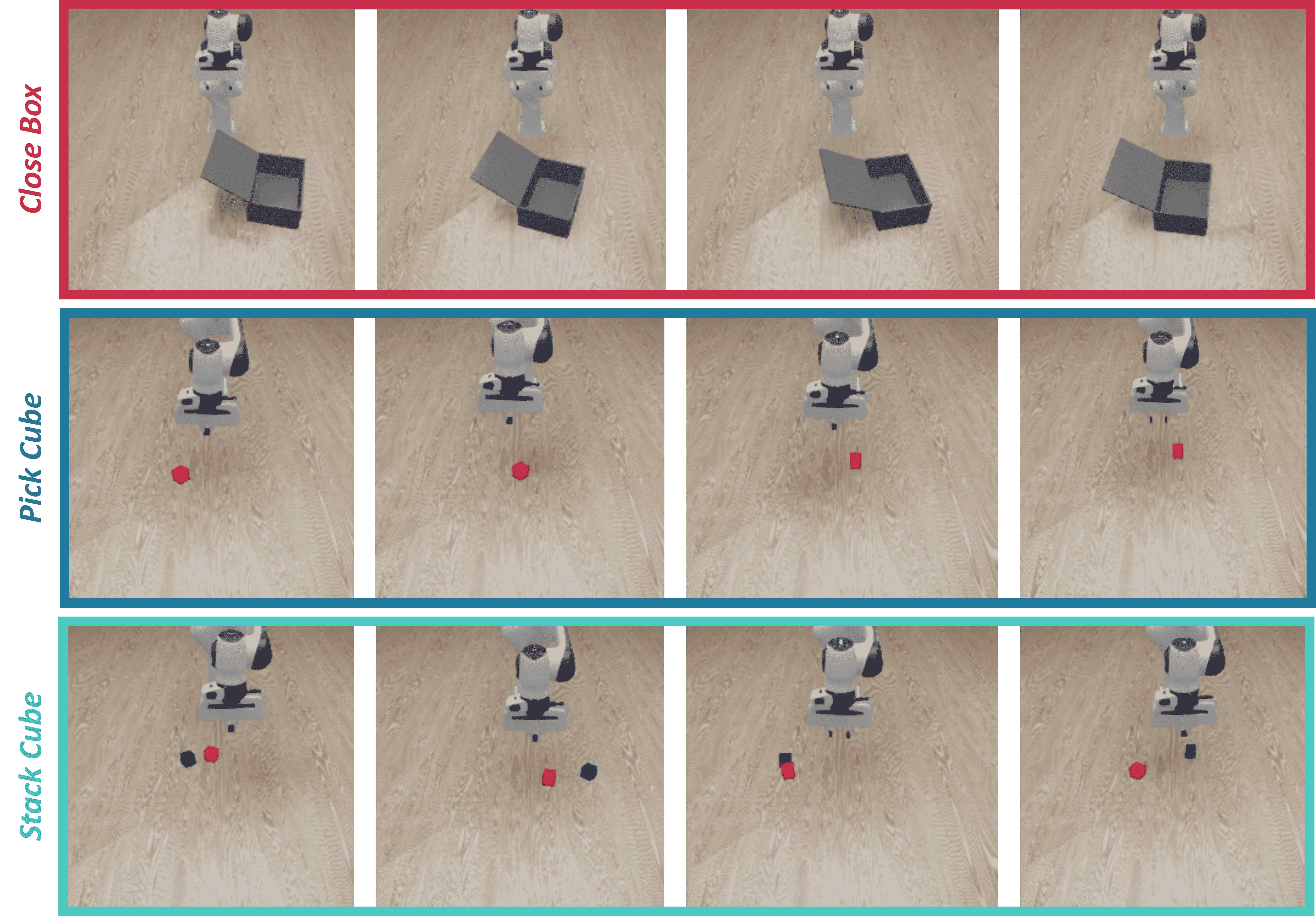}
		\caption{\textbf{Randomization Level 0.} Level 0 introduces the randomization of initial object positions. It serves as the training set.}
		\label{level0}
	\end{figure*}
	
	\renewcommand*{\thefigure}{S5}
	\begin{figure*}
		\centering
		\includegraphics[width=0.7\linewidth]{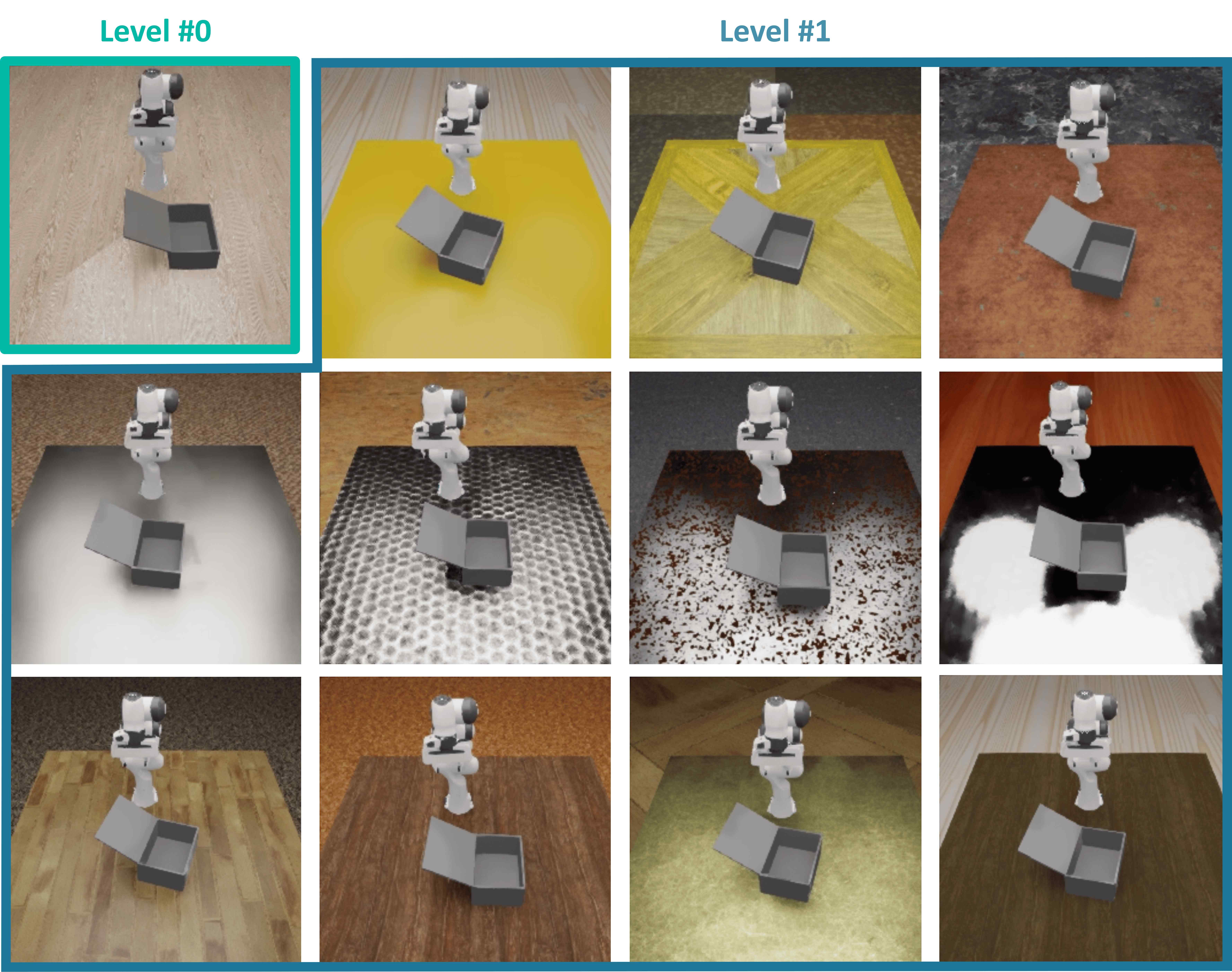}
		\caption{\textbf{Randomization Level 1.} Level 1 primarily focuses on the randomization of the environmental background.}
		\label{level1}
	\end{figure*}
	
	\renewcommand*{\thefigure}{S6}
	\begin{figure*}
		\centering
		\includegraphics[width=0.9\linewidth]{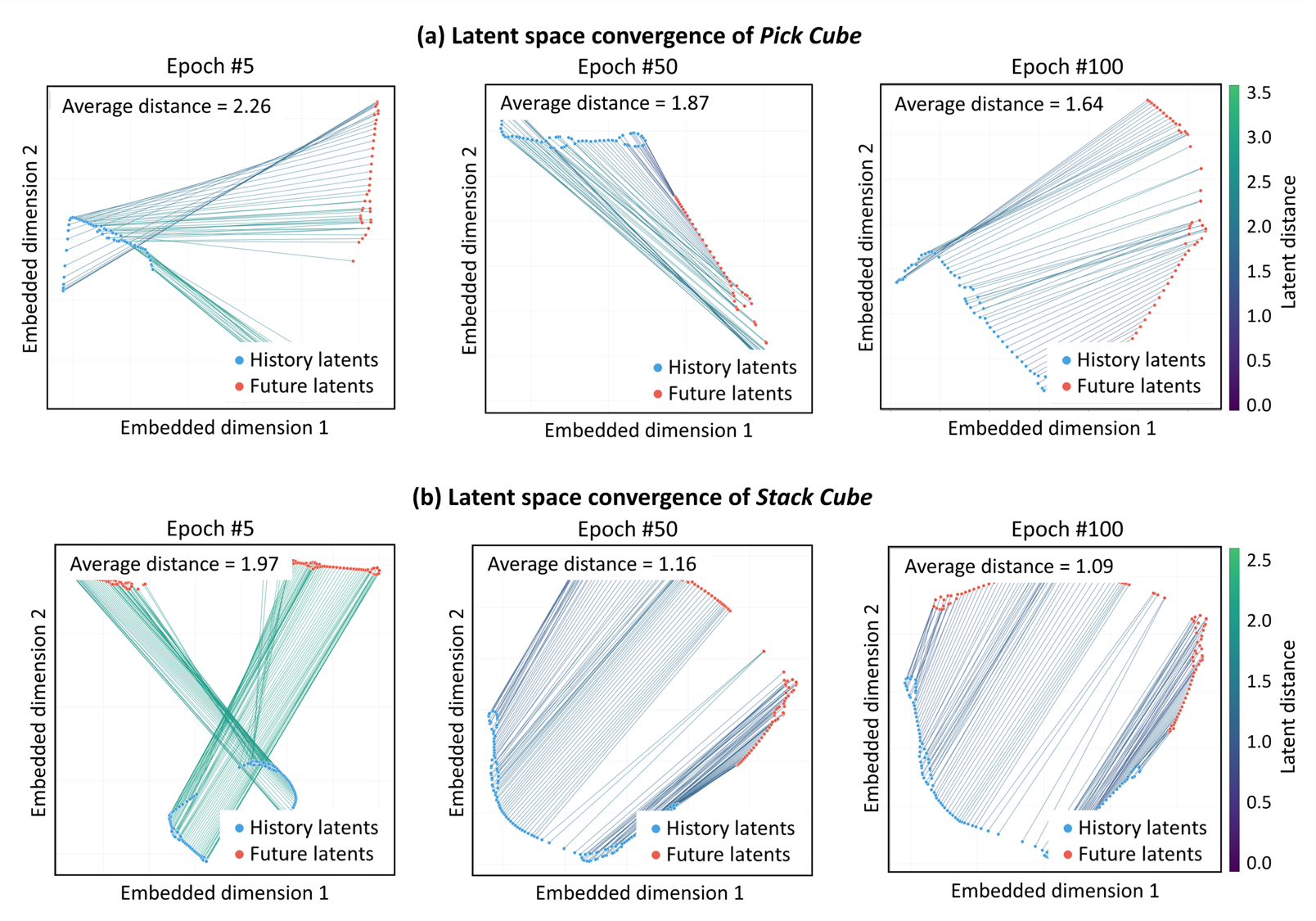}
		\caption{\textbf{Convergence of latent space representations in \textit{Pick Cube} and \textit{Stack Cube} tasks.} We apply t-SNE to jointly embed history and future action latents, with paired samples connected by lines. Line colors indicate distances computed in the 512-dimensional latent space.}
		\label{latent_space_appen}
	\end{figure*}

	\renewcommand*{\thefigure}{S7}
	\begin{figure*}
		\centering
		\includegraphics[width=0.8\linewidth]{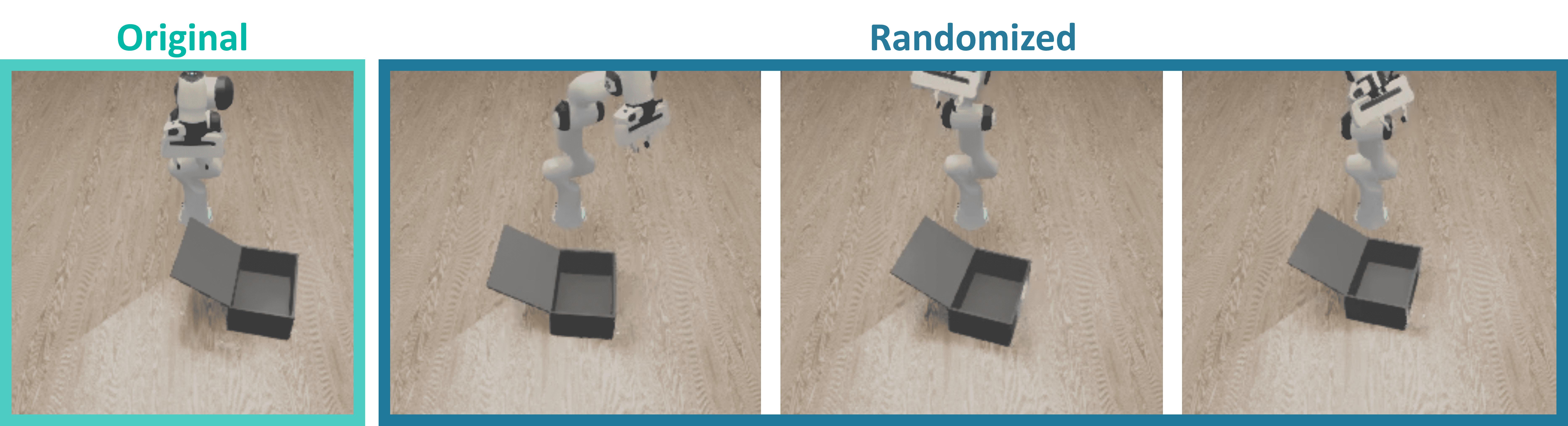}
		\caption{\textbf{Initial state uncertainty.} We randomize the initial pose of the robot to investigate the generalization performance under action-level uncertainty.}
		\label{initial_state}
	\end{figure*}
	
	\renewcommand*{\thefigure}{S8}
	\begin{figure*}
		\centering
		\includegraphics[width=0.5\linewidth]{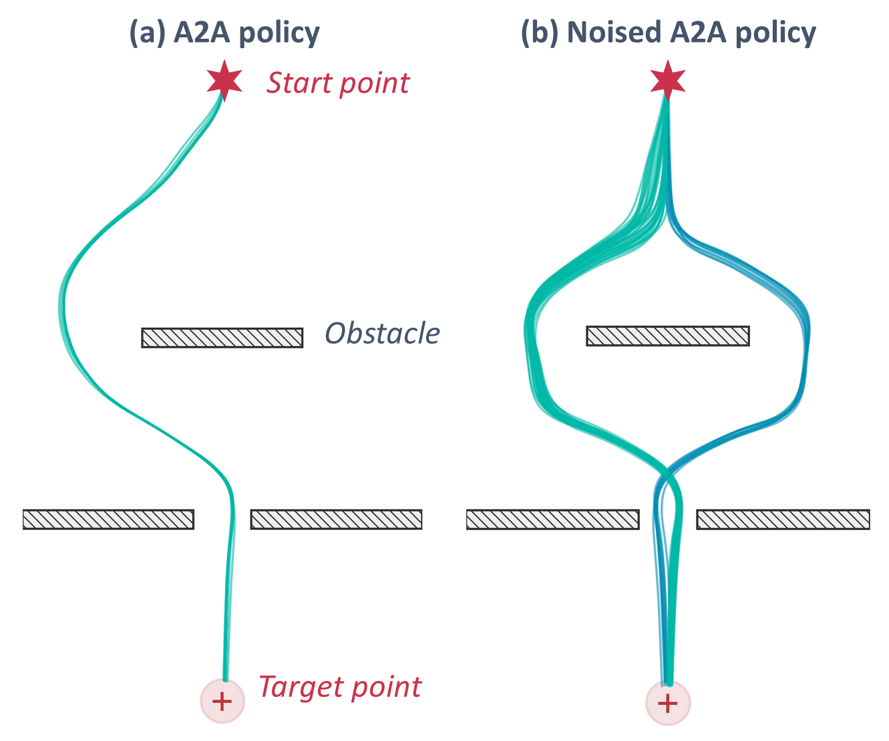}
		\caption{\textbf{Subtle action noise unlocks multimodal behavior.} 
			A 2D navigation task in which the agent must reach the target (\textcolor{red}{$\oplus$}) from the start (\textcolor{red}{$\star$}) by passing through a gap between obstacles, a setting with inherently multimodal optimal solutions (going left or right of the upper obstacle). 
			{(a)} The A2A policy is effectively deterministic: rollouts collapse to a single mode despite the multimodal training data. 
			{(b)} Adding zero-mean \textit{Gaussian} noise with std $\sigma=0.02$ to the historical actions $\mathbf{a}_{\leq t}$ breaks this determinism without sacrificing task success, allowing the policy to express both modes across rollouts. 
			This validates our design choice of action perturbation as a lightweight mechanism for preserving the multimodality of the learned policy.}
		\label{multi_modal}
	\end{figure*}

\end{document}

%% file: sec/intro.tex
\section{Introduction}


Recent advances in imitation learning architectures have significantly expanded the capabilities of robotic systems operating in complex and unstructured environments~\citep{ai2025review}. Among these, diffusion-based policies~\citep{ddpm, flowmatching, dp, intelligence2025pi} have emerged as a powerful paradigm for modeling the intrinsic multi-modality of human demonstrations. These methods formulate action generation as a conditional denoising process: during training, a neural network learns to predict injected noise, while at inference time, executable actions are obtained by iteratively denoising samples initialized from random noise.


Despite their strong empirical performance on high-precision and multi-modal tasks, diffusion models suffer from a well-known limitation: the ``denoise-from-scratch'' paradigm incurs substantial inference latency~\citep{pan2025much}. Generating a single action typically requires dozens of iterative denoising steps, creating a major bottleneck for real-time robotic control, where low cycle time and rapid feedback are essential for stable execution.
To mitigate this limitation, prior work has explored improving the initialization of the diffusion process to accelerate inference. For instance,~\citet{steeringdp} propose an RL-trained policy to steer the initial sampling distribution, while~\citet{warmstarts} employ warm-start strategies to identify more informative starting points. These approaches replace uninformed \textit{Gaussian} noise with priors closer to the data distribution, often through auxiliary models that predict distributional statistics conditioned on the current observation. While effective, such approaches still rely on stochastic noise initialization and inevitably introduce additional modeling complexity.

This observation raises a more fundamental question: \textbf{do robotic policies truly need to be generated by sampling from random noise?} Diffusion models were originally developed for high-fidelity image synthesis and video generation~\citep{ddpm,song2020denoising,rombach2022high,flowmatching,rectifiedflow,blattmann2023stable}, where generation typically begins from uninformed noise due to the absence of meaningful priors. Robot control, however, operates under a fundamentally different regime. Modern robots are equipped with rich state sensors that provide continuous, low-latency feedback about the system’s configuration and dynamics~\citep{jia2024feedback, jia2024learning}. This structured feedback constitutes a strong and reliable prior, naturally encoding the robot’s current physical state or recent execution history, and thus offers a principled alternative to random noise initialization for action generation~\citep{jia2025foreseer}.


However, as shown in Fig.~\ref{framework} (a) and (b), most existing approaches adhere to either regression- or diffusion-based paradigms that condition action generation on a simple concatenation of the current proprioceptive state/action and visual observations. Such designs overlook the temporal continuity intrinsic to robotic motion and often dilute low-dimensional proprioceptive signals when fused directly with high-dimensional visual representations. Moreover, recent studies show that explicit conditioning on proprioceptive states can adversely affect spatial generalization~\citep{zhao2025you}.
Consequently, rich historical information about system dynamics and action trends remains largely underexploited in prevailing frameworks. Notably, diffusion models define mappings between probability distributions rather than individual states. This perspective motivates a natural question: can the inherent proximity between distributions of past executions and future actions be exploited to reduce learning complexity, yielding a shorter and more stable transport path for policy generation?


To this end, we propose \textbf{A}ction-\textbf{to}-\textbf{A}ction {Flow} Matching (\textbf{\m{}}), a novel flowing matching-based policy paradigm that shifts action generation \textit{from uninformed sampling to informed initialization}. Unlike prior approaches that initialize the diffusion process with standard \textit{Gaussian} noise, \m{} directly leverages a sequence of historical proprioceptive actions as the starting point for action generation, as illustrated in Fig.~\ref{framework}. To capture subtle motion patterns and temporal dependencies, these low-dimensional action histories are embedded into a high-dimensional latent space. By learning a flow that transports historical action distributions to future actions, \m{} bypasses the costly iterative denoising process from \textit{Gaussian} noise.
We evaluate \m{} extensively in both simulated environments and real-world robotic systems. Empirically, our method exhibits remarkable training efficiency and consistently outperforms 8 state-of-the-art baselines.
In particular, \m{} achieves significantly faster inference convergence, i.e., up to 20$\times$ and $5\times$ faster than vanilla diffusion and flow matching methods, respectively, while \textbf{enabling high-quality action generation in as few as a single inference step}. 
Moreover, grounding the generation process in proprioceptive history substantially improves robustness to visual perturbations, and the history-informed initialization enhances generalization to unseen configurations by enforcing physical consistency over time.

In summary, we introduce a new diffusion-based robot policy paradigm that replaces stochastic noise initialization with history-based proprioceptive initialization, enabling informed action generation grounded in the robot’s own dynamics. Leveraging latent state representations, \m{} captures fine-grained motion structure and supports efficient action-to-action transitions without iterative denoising from random noise. Extensive experiments in both simulated and real-world robotic environments demonstrate that our method achieves state-of-the-art performance across training efficiency, inference speed, robustness to visual perturbations, and generalization to unseen configurations. Furthermore, we showcase the applicability of \m{} in robotic video generation, suggesting promising potential for broader scalability.

%% file: sec/related.tex
\section{Related work}

\subsection{Visuomotor policy}
Visuomotor policy is a robot learning framework that maps raw sensory observations, typically high-dimensional visual inputs and low-dimensional robotic states, directly into low-level control actions. Early approaches, such as action chunking (ACT)~\citep{zhao2023learning}, utilized a conditional variational autoencoder with Transformers~\citep{vaswani2017attention} to learn fine-grained bimanual manipulation by predicting future action sequences. Diffusion policy~\citep{dp} introduced a diffusion generative paradigm by modeling the action distribution as a score-based gradient field~\citep{ddpm, song2020score}, significantly improving stability in multi-modal environments. Flow matching \citep{flowmatching} used in Vision-Language-Action (VLA) models, like $\pi$ family~\citep{intelligence2025pi,intelligence2025pi05, black2024pi_0}, has sought to simplify the generative process by learning straight-line probability paths between noise and action distributions. More recently, \citet{pan2025much} suggest that the selection of advanced model architectures (e.g., DiT~\citep{Peebles_2023_ICCV} and UNet\citep{dp}) and action chunking strategy~\citep{zhao2023learning} has a more significant impact on the success of flow matching than the regression method.

Despite these successes, diffusion methods suffer from high computational costs due to their iterative multi-step inference nature and complex architectures. Conversely, the vision-to-action model (VITA)~\citep{gao2025vita} pursues architectural minimalism by employing a lightweight Multi-Layer Perceptron (MLP) backbone for direct vision-to-action mapping. However, VITA relies completely on visual inputs, making it vulnerable to environmental visual distractors. Furthermore, VITA exhibits a dimensionality mismatch within its flow space. Recent efforts such as MeanFlow-based policies~\citep{fang2025omp, geng2025mean} and consistency models~\citep{zhang2025flowpolicy,li2026ofp, Consistency} have begun to push one-step generation into robotic action prediction by optimizing the theoretical boundaries of the noise-to-action mapping. In contrast, \m{} exploits the physical continuity of actions to directly map proprioceptive historical actions to future actions, fundamentally shortening the generative path and enabling high-performance single-step generation.

\subsection{Noise optimization of diffusion}

In image and video synthesis, optimizing initial noise is a key strategy for enhancing generation quality and speed~\citep{ahn2024noise,mao2024lottery,samuel2024generating,eyring2024reno, warmstarts}. For example, \citet{ahn2024noise} eliminate the need for computationally expensive guidance techniques by learning to map standard \textit{Gaussian} noise to a guidance-free noise space through a lightweight LoRA module~\citep{hu2022lora}. Similarly, warm-start diffusion~\citep{warmstarts} utilizes a deterministic model to provide an informed mean and variance of the initial \textit{Gaussian} distribution based on context, reducing the sampling path. In robotics, however, noise optimization is rarely explored. \citet{steeringdp} adapts behavioral cloning policies by running reinforcement learning over the latent-noise space, allowing the agent to steer robot actions without altering the base policy weights.

Departing from the long transport paths of the noise-to-data paradigm, \citet{chen2024bridger} leverage informative initial distributions, yet remains constrained by its dependence on heuristic or pre-trained models. Our \m{} introduces an action-to-action transport mechanism. By directly employing clean historical data as the initial distribution, we leverage the physical continuity of robotic motion to place the starting point much closer to the target in the high-dimensional latent space. This reduced distributional gap bypasses the need for multi-step refinement, enabling high-fidelity single-step inference optimized for real-time robotic control. 

%% file: sec/conclusion.tex


\section{Conclusion}
This paper introduces \m{}, an efficient generative paradigm that replaces noise-based initialization with action-to-action transport. By leveraging the physical consistency of sequential motions, \m{} aligns starting and target distributions, enabling a lightweight MLP to achieve high success rates with minimal latency. Across diverse simulation and real-world robotic benchmarks, \m{} matches or surpasses state-of-the-art diffusion and flow-based policies while reducing inference to fewer steps. This approach effectively eliminates the computational bottlenecks typical of diffusion-based policies. Beyond robotics and video generation, the framework is inherently suited for diverse continuous temporal tasks, and its potential in domains characterized by sequential continuity remains a promising avenue for future exploration.


\section{Limitation}
\m{} is grounded in the physical continuity of actions, which makes it most effective for tasks dominated by smooth, continuous control signals. For discrete or switch-like action dimensions, such as binary gripper open/close commands, this continuity prior provides limited benefit. Moreover, the current objective requires manual tuning of several loss-weighting coefficients. We leave adaptive loss weighting and support for hybrid continuous-discrete action spaces as promising directions for future work.